\documentclass[accepted]{uai2026} 
                        
\usepackage[american]{babel}

\usepackage{natbib} 
\usepackage{mathtools} 
\usepackage{booktabs} 
\usepackage{tikz} 

\usepackage{import}
\usepackage{amsthm}
\usepackage{url}
\usepackage{xcolor}
\usepackage{wrapfig}
\usepackage{float}
\usepackage{algorithm}
\usepackage[noend]{algpseudocode}
\algnewcommand{\LineComment}[1]{\State \(\triangleright\) #1}
\algnewcommand\algorithmicforeach{\textbf{for each}}
\algdef{S}[FOR]{ForEach}[1]{\algorithmicforeach\ #1\ \algorithmicdo}
\algrenewcommand\algorithmicthen{}
\usepackage{makecell}
\usepackage{multirow}
\usepackage{subcaption}
\usepackage{caption}
\usepackage{cleveref}

\newtheorem{theorem}{Theorem}
\newtheorem{corollary}{Corollary}
\newtheorem{lemma}{Lemma}
\newtheorem{definition}{Definition}

\newtheorem{proposition}{Proposition}

\def\Pr{\mathop{\rm Pr}\nolimits}

\def\CC{{\cal C}}

\def\X{{\mathbf X}}
\def\Y{{\mathbf Y}}
\def\Z{{\mathbf Z}}
\def\U{{\mathbf U}}
\def\W{{\mathbf W}}

\def\V{{\mathbf V}}
\def\P{{\mathbf P}}

\def\S{{\mathbf S}}

\def\H{{\mathbf H}}
\def\x{{\mathbf x}}
\def\y{{\mathbf y}}
\def\D{{\mathbf D}}

\def\u{{\mathbf u}}
\def\v{{\mathbf v}}

\def\p{{\mathbf p}}
\def\s{{\mathbf s}}

\newcommand{\indep}{\!\perp\!\!\!\perp}

\newcommand{\C}{\mathbf C}

\usepackage{tikz}
\usetikzlibrary{arrows}
\tikzset{
state/.style = {shape=rectangle,draw,font=\huge},
sstate/.style = {draw,thick,minimum size=3.0em,font=\huge},
dstate/.style = {shape=circle,draw,thick,double,minimum size=3.0em,font=\huge},
square/.style={regular polygon,regular polygon sides=4},
bidirected/.style={Latex-Latex,dashed},
}
\usetikzlibrary{shapes,decorations,calc,fit,positioning}
\usetikzlibrary{arrows.meta}
\tikzset{>={Latex[width=1.5mm,length=1.5mm]}}

\newcommand\dagsize{0.35}

\def\Equal{\texttt{=}}

\title{On the Granularity of Causal Effect Identifiability}

\author[1]{\href{mailto:<yizuo.chen@ucla.edu>}{Yizuo Chen}{}}
\author[1]{\href{mailto:<darwiche@cs.ucla.edu>}{Adnan Darwiche}{}}

\affil[1]{%
    Computer Science Department\\
    University of California, Los Angeles\\
    USA
}
  
\begin{document}
\maketitle

\begin{abstract}
The classical notion of causal effect identifiability is defined in terms of treatment and outcome variables. In this paper, we consider the identifiability of state-based causal effects: how an intervention on a particular \emph{state} of treatment variables affects a particular \emph{state} of outcome variables. We demonstrate that state-based causal effects may be identifiable even when variable-based causal effects may not. Moreover, we show that this separation occurs only when additional knowledge --- such as context-specific independencies --- is available. We further examine knowledge that constrains the states of variables, and show that such knowledge can improve both variable-based and state-based identifiability when combined with other knowledge such as context-specific independencies. We finally propose an approach for identifying causal effects under these additional constraints, and conduct empirical studies to further illustrate the separations between the two levels of identifiability. 
\end{abstract}

\section{Introduction}
The causal effect is a distribution on some outcome variables that materializes as a result of an intervention on a set of treatment variables;
e.g., the probability that a patient will recover if the doctor instructs them to take a drug. In general, a causal effect cannot be estimated based on observational data only, therefore requiring experimental studies; see, e.g.,~\citep{pearl18}.
However, 
given a graph that encodes the causal relations among variables, a causal effect may in some cases be uniquely determined (i.e., estimable) using observational data only. In such cases, the causal effect is said to be \emph{identifiable} and a formula may be derived to estimate it based on observational data. 

Both complete and incomplete methods have been proposed in the past for testing the identifiability of causal effects based on causal graphs; see, e.g.,~\citep{pearl00b, hernanbook} for some overviews. Some early methods for testing identifiability include the back-door criterion~\citep{pearl1993b,pearl00b} and front-door criterion~\citep{pearl95,pearl00b}. These methods are sound yet not complete and hence may fail to identify causal effects that are indeed identifiable. Complete methods for identifying causal effects include the do-calculus~\citep{pearl00b}, the identification algorithm in~\citep{TianPearl03,aaai/HuangV06}, and the ID algorithm in~\citep{aaai/ShpitserP06,jmlr/ShpitserP08}. 
All these methods, however, are \emph{variable-based}
in that they test
whether the causal effect is identifiable for \textit{every} possible intervention on the treatment variables and \textit{every} possible state of the outcome variables (for the same set of treatment and outcome variables).
We will argue in this paper that variable-based identifiability is appropriate when the 
available information consists only of a causal graph and some observational
data but can be too coarse-grained when additional information
is available. In particular, given additional information, it is conceivable that the causal effect of \textit{some} treatment on \textit{some} outcome may be identifiable, while the causal effect of some other treatment on some other outcome may not be identifiable. This highlights the need for more refined (state-based) approaches to identifiability as we discuss and present later.

This distinction becomes particularly relevant in light of 
recent works that study the identifiability problem
while assuming additional knowledge beyond causal graphs and observational data. This includes knowledge of functional dependencies~\citep{neurips/ChenDarwiche24}, context-specific independencies (CSIs)~\citep{nips/TikkaHK19,aistats/MokhtarianJEK22}, and known parameters~\citep{aict/ChenDarwiche25}.
Such additional knowledge can be formulated as constraints, leading to the notion of \emph{constrained-identifiability}~\citep{neurips/ChenDarwiche24,aict/ChenDarwiche25}. The main use of such additional knowledge has been to show that it may render some (variable-based) causal effects identifiable which would otherwise be unidentifiable without the additional knowledge. We show in this paper that such knowledge has an additional value: it renders the distinction between variable-based causal effects and state-based causal effects meaningful and worthy of study
as it highlights some new directions for improving causal effect identifiability. 

Let $\X$ be the treatment variables and $\Y$ be the outcome variables. We start 
in Section~\ref{sec:two-levels} by formulating state-based identifiability which is concerned with whether the causal effect of a particular treatment ($\X\!\!=\!\!\x$) on a particular outcome ($\Y\!\!=\!\!\y$) can be uniquely determined based on available information. We then illustrate in Section~\ref{sec:csi} the distinction between state-based and variable-based identifiability given knowledge in the form of context-specific independencies (CSIs), which are effectively constraints on model parameters that cannot be captured by causal graphs.
We follow in Section~\ref{sec:domain-constraints} 
by considering state constraints as an additional 
type of knowledge, showing that they are not useful on their own in the classical identifiability setup, but they may widen the gap between state-based and variable-based identifiability when coupled with other knowledge such as CSIs. We then present a novel method in Section~\ref{sec:main-algs} 
for identifying both variable-based and state-based causal effects under CSIs and state constraints, and follow by empirically evaluating its completeness and computationally efficiency in Section~\ref{sec:exp}. We finally close with some concluding remarks in Section~\ref{sec:conclusion}.
All proofs are included in Appendix~\ref{app:proof}.

\section{Two Levels of Causal Effect Identifiability}
\label{sec:two-levels}
All variables are assumed to be discrete throughout this work. Let \(G\) denote a causal graph in the form of a DAG, \(\V\) the set of observed variables in $G$, \(\X\) the treatment variables, and \(\Y\) the outcome variables. Variables outside \(\V\) are called \emph{hidden} and are denoted by \(\U.\) Let \(\x\) and \(\y\) denote some states (instantiations) of variables \(\X\) and \(\Y,\) respectively.

A parameterization (model) of causal graph $G$ assigns a conditional probability table (CPT) \(\theta_{X | \P}\) to each variable \(X\) with parents \(\P\) in \(G\), representing the conditional distribution \(\Pr(X | \P).\) Such a model induces the distribution
$\Pr(\V,\U) = \prod_{X \in \V \cup \U}\theta_{X | \P}$ and the
\textit{observational distribution} \(\Pr(\V) = \sum_\U \Pr(\V,\U).\) 
The probability of \(\y\) under an intervention \(\x\) is denoted \(\Pr_\x(\y)\)
and computed using the \emph{interventional distribution} \(\Pr_\x(\V, \U)\) defined as follows; see~\cite[Ch.~3]{pearl00b}.
We first remove the incoming edges of \(\X\) in $G$ and replace the CPT for each \(X \in \X\) by a new CPT \(\theta_X\), where \(\theta_X(x) = 1\) if \(x \in \x\) and \(\theta_X(x) = 0\) otherwise. The updated set of CPTs induces the interventional distribution \(\Pr_\x(\V, \U)\).
The classical identifiability problem checks whether any two models 
(parameterizations of \(G\)) that induce the same observational distribution \(\Pr(\V)\) also yield the same causal effect \(\Pr_\x(\y)\) for all $\x$ and all $\y.$

\begin{definition}[Classical Identifiability]
\label{def:classical-id}
The causal effect of \(\X\) on \(\Y\), denoted \(\Pr_\X(\Y),\) is identifiable wrt \(\langle  G, \V \rangle\) if \(\Pr^1_\x(\y) = \Pr^2_\x(\y)\) for all instantiations \(\x, \y\) and any pair of models that induce \(\Pr^1, \Pr^2\) such that \(\Pr^1(\V) = \Pr^2(\V).\)\footnote{We assume that the observational distribution \(\Pr(\V)\) is strictly positive, i.e., \(\Pr(\V) > 0.\) This positivity assumption is common in existing identification methods such as \textsc{identify}~\citep{TianPearl03} and the do-calculus~\citep{pearl00b}.}
\end{definition}

Classical identifiability is variable-based in the following sense. If the causal effect $\Pr_\X(\Y)$ is identifiable, then $\Pr_\x(\y)$ is uniquely computable from the observational distribution $\Pr(\V)$
for \emph{every} possible treatment $\x$ and \emph{every} possible outcome $\y.$ 
Note that classical identifiability does not restrict the states of model variables.

We now introduce a more granular notion of identifiability which targets a particular treatment $\x$ and a particular outcome $\y.$ For example, the probability that a major accident will occur if a particular speed limit is enforced, 
\(\Pr_{\mathsf{speedLimit}=65}(\mathsf{accident}\!=\!\textit{major}).\)

\begin{definition}[State-Based Identifiability]
\label{def:inst-id}
The causal effect of \(\x\) on \(\y,\) denoted \(\Pr_\x(\y),\) is identifiable wrt \(\langle  G, \V \rangle\) if \(\Pr^1_\x(\y) = \Pr^2_\x(\y)\) for any pair of models that induce \(\Pr^1, \Pr^2\) such that \(\Pr^1(\V) = \Pr^2(\V).\)
\end{definition}

State-based identifiability refers to a particular treatment $\x$ and a particular outcome $\y$ which are states of the treatment and outcome variables. To illustrate the applicability of state-based identifiability, consider the following real-world scenarios. Suppose that speed limit signs (e.g., a maximum of 65 mph) have been installed on \emph{some} streets in a district, and the policymakers are interested in the following question: how much would enforcing speed limits on \emph{all} streets reduce the accident rate? This question requires comparing the state-based causal effect $\Pr_{\mathsf{SpeedLimit}\Equal\textit{true}}(\mathsf{Accident}\Equal\textit{true})$ with the observational probability $\Pr(\mathsf{Accident}\Equal\textit{true})$. The policymakers are typically not interested in the accident rate under a policy that removes all speed limit signs, i.e., $\Pr_{\mathsf{SpeedLimit}\Equal\textit{false}}(\mathsf{Accident}\Equal\textit{true})$ --- a quantity that would be computed in the variable-based causal effect $\Pr_{\mathsf{SpeedLimit}}(\mathsf{Accident})$ but is irrelevant in this scenario.

As another example, suppose a hospital is concerned with patients' recovery rates from a disease as well as the side effects of taking a drug. To decide whether it is beneficial to enforce \emph{all} patients to take the drug, the hospital may be interested in the state-based causal effect $\Pr_{\mathsf{Drug}\Equal\textit{true}}(\mathsf{Recover}\Equal\textit{true}, \mathsf{SideEffects}\Equal\textit{low})$, which is easier to obtain than the variable-based causal effect \(\Pr_{\mathsf{Drug}}(\mathsf{Recover}, \mathsf{SideEffects})\) commonly considered for evaluating the drug's therapeutic effect.
These two examples show that state-based causal effects become relevant when the focus is on the effect of an intervention relative to the current system (captured by the observational distribution), rather than on comparing the effects of distinct interventions as typically done in the classical setup.

The distinction between variable-based and state-based identifiability is not interesting if the only information available consists of a causal graph $G$ and an observational distribution $\Pr(\V).$ In fact, these two notions are equivalent in the classical setup, which we formulate as the following proposition and include a proof in Appendix~\ref{app:proof}.
\begin{proposition}
\label{prop:eq-classical}
The causal effect \(\Pr_\X(\Y)\) is identifiable iff \(\Pr_\x(\y)\) is identifiable for some states \(\x, \y.\)
\end{proposition}

The distinction between the two notions materializes though if we have additional knowledge beyond $G$ and $\Pr(\V).$ We elaborate on this next.

\section{The Impact of CSI Constraints on Identifiability}
\label{sec:csi}
We next show how CSI constraints~\citep{uai/BoutilierFGK96,pgm/ShenCD20}, studied for causal effect identifiability in~\citep{nips/TikkaHK19,aistats/MokhtarianJEK22,aict/ChenDarwiche25}, draw a distinction between variable-based and state-based identifiability.
We consider CSIs in the form of (\(Y \indep P_1 | \p_2, \P_3\)), where 
$Y$ is a node in the causal graph and 
\(\{P_1\}, \P_2, \P_3\) is a partition of its parents. This CSI constraint is
interpreted as follows: \(Y\) is independent of \(P_1\) under the specific state \(\p_2\) of $\P_2$ and any states of \(\P_3.\) We call \(\P_2\) the \emph{context variables} and \(\p_2\) the \emph{context} of the CSI.
For example, the CSI (\(\mathsf{Light} \indep \mathsf{People}\ |\ \mathsf{Outage}\Equal \textit{yes}, \mathsf{Time}\)) states that 
when a power outage occurs (context), the status of light becomes independent of the presence of people, regardless of the time of day. This independence, however, does not hold when the power outage does not occur. CSI constraints are effectively constraints on model parameters so they cannot be captured using the language of causal graphs. Constraints, such as CSIs, restrict the set of models
considered when deciding identifiability.
This has the impact of making some causal effects identifiable, which would be unidentifiable without the constraints. We assume that the constraints discussed in this work are known and provided as additional inputs to the identifiability problem. Our results are therefore independent of how these constraints are obtained, whether specified from domain knowledge or learned from data. To continue with this discussion, we need to recall the following definition of constrained identifiability (at both the variable-based and state-based levels).

\begin{definition}\citep{aict/ChenDarwiche25}
\label{def:constrained-id}
Let \(\CC\) be a set of constraints.
A causal effect \(\Pr_\x(\y)\) is \underline{identifiable} wrt \(\langle  G, \V, \CC \rangle\) if \(\Pr^1_\x(\y) = \Pr^2_\x(\y)\) for distributions  \(\Pr^1, \Pr^2\) that are induced by $G$, satisfy constraints $\CC$ and \(\Pr^1(\V) = \Pr^2(\V).\) Moreover, the causal effect \(\Pr_\X(\Y)\) is \underline{identifiable} wrt \(\langle  G, \V, \CC \rangle\) if \(\Pr_\x(\y)\) is identifiable for all \(\x,\y.\)
\end{definition}

Our first result states that the (state-based) identifiability of \(\Pr_\x(\y)\) and the (variable-based) identifiability of \(\Pr_\X(\Y)\) are equivalent if none of the states \(\x,\y\) appear in the contexts of CSIs. For example, if (\(\mathsf{Light} \indep \mathsf{People}\ |\ \mathsf{Outage}\Equal \textit{yes}, \mathsf{Time})\) is the only CSI constraint, then \(\Pr_{\mathsf{Outage}\Equal \textit{no}}(\mathsf{Light}\Equal \textit{off})\) is identifiable iff \(\Pr_{\mathsf{Outage}}(\mathsf{Light})\) is identifiable. The result is stated in the following proposition.
\begin{proposition}
\label{prop:state-var-eq}
Consider Definition~\ref{def:constrained-id} of constrained identifiability and let \(\CC\) be a set of CSIs. If none of the states \(\x,\y\) appear in the contexts of \(\CC,\) then \(\Pr_\x(\y)\) is identifiable wrt \(\langle G, \V, \CC \rangle\) iff \(\Pr_\X(\Y)\) is identifiable wrt \(\langle G, \V, \CC \rangle\).
\end{proposition}

It immediately follows that the (variable-based) identifiability of \(\Pr_\X(\Y)\) can be reduced to the (state-based) identifiability of \(\Pr_\x(\y)\) for states \(\x,\y\) that do not appear in CSI contexts. Hence, if we have an algorithm for testing state-based identifiability under CSIs, we can use it to test variable-based identifiability under CSIs by choosing states \(\x,\y\) that do not appear in CSI contexts. This can always be done as we are only considering CSI constraints and these do not constrain variable states.

However, in general, state-based and variable-based identifiability are not equivalent. We demonstrate the separation between the two using the following example. Consider the causal graph in Figure~\ref{sfig:csi-eg1} where $\mathsf{Age}$ and $\mathsf{Degree}$ are hidden, and 
$\mathsf{Years}$ refers to years of experience. 
Suppose now that the company passes a policy that ignores the employee's degree when they decide the salary for entry-level employees, which gives the CSI (\(S \indep D\ |\ \) \( Y, J \Equal \textit{entry-level}\)).
The state-based causal effect \(\Pr_{Y \geq 10}(J \Equal \textit{entry-level}, S\Equal \textit{low})\) is now identifiable;
see Proposition~\ref{prop:cid-diff} in Appendix~\ref{app:proof}. However, a different state-based causal effect \(\Pr_{Y \geq 10}(J \Equal \textit{senior-level}, S \Equal \textit{low})\) is unidentifiable since we cannot leverage the CSI when \(J \Equal \textit{senior-level}\) (the CSI only takes effect when \(J \Equal \textit{entry-level}\)). Therefore, the variable-based causal effect \(\Pr_Y(J, S)\) is unidentifiable yet the state-based causal effect \(\Pr_{Y \geq 10}(J \Equal \textit{entry-level}, S\Equal \textit{low})\) is identifiable. 

This example shows that an unidentifiable causal effect at the variable level may be identifiable for a specific treatment and outcome. Hence, existing methods for variable-based identifiability are no longer sufficient for state-based identifiability, highlighting the need for more refined approaches to identify causal effects at this more granular level.

The next question we consider is whether CSI constraints can improve variable-based identifiability like they do for state-based identifiability. That is, if a causal effect is unidentifiable (according to Definition~\ref{def:classical-id}), can it become identifiable when CSI constraints are imposed (according to Definition~\ref{def:constrained-id})? The answer turns out to be positive. Consider the modified causal graph for the company example depicted in Figure~\ref{sfig:csi-eg2}, where \(Y\) is no longer a parent of \(S\). In this graph, the variable-based causal effect \(\Pr_{J}(S)\) is unidentifiable. However, suppose we further know that the company determines an employee's job level only based on their age if they have worked for over 10 years --- i.e., we have the CSI \((J \indep D\ |\ A, Y\geq 10)\). The same causal effect \(\Pr_{J}(S)\) becomes identifiable in this case and can be estimated as \(\Pr_J(S) = \Pr(S | J, Y\geq 10).\) We formalize this result as Proposition~\ref{prop:csi-var-useful} in Appendix~\ref{app:proof}. The result suggests that classical identification methods (such as \textsc{identify}~\citep{TianPearl03}) are insufficient for variable-based identifiability when additional CSIs are available.

\begin{figure}[tb]
\centering
\begin{subfigure}[b]{0.49\linewidth}
\centering
\begin{tikzpicture}[->=stealth,auto,scale=\dagsize,transform shape]
\node[rectangle,draw,font=\huge] (age) at (0,0) {Age ($A$)};
\node[font=\huge] (years) at (-2,-2) {Years ($Y$)};
\node[rectangle,draw,font=\huge] (degree) at (4,0) {Degree ($D$)};
\node[font=\huge] (level) at (2,-2) {Job ($J$)};
\node[rectangle,font=\huge] (salary) at (6,-2) {Salary ($S$)};
\path (age) edge (years);
\path (age) edge (level);
\path (degree) edge (level);
\path (degree) edge (salary);
\path (level) edge (salary);
\path (years) edge[bend right=10] (salary);
\end{tikzpicture}
\caption{}
\label{sfig:csi-eg1}
\end{subfigure}
\begin{subfigure}[b]{0.49\linewidth}
\centering
\begin{tikzpicture}[->=stealth,auto,scale=\dagsize,transform shape]
\node[rectangle,draw,font=\huge] (age) at (0,0) {Age ($A$)};
\node[font=\huge] (years) at (-2,-2) {Years ($Y$)};
\node[rectangle,draw,font=\huge] (degree) at (4,0) {Degree ($D$)};
\node[font=\huge] (level) at (2,-2) {Job ($J$)};
\node[rectangle,font=\huge] (salary) at (6,-2) {Salary ($S$)};
\path (age) edge (years);
\path (age) edge (level);
\path (degree) edge (level);
\path (degree) edge (salary);
\path (level) edge (salary);
\path (years) edge (level);
\end{tikzpicture}
\caption{}
\label{sfig:csi-eg2}
\end{subfigure}
\caption{causal graphs showing the effectiveness of CSI constraints, with hidden variables shown in rectangles.}
\label{fig:csi-eg}
\end{figure} 

\section{The Impact of State Constraints on Identifiability}
\label{sec:domain-constraints}
We showed earlier that CSI constraints alone can improve both variable-based and state-based identifiability.
We next show that these constraints can further improve identifiability when coupled with \emph{state constraints,} which specify the possible states of some variables.
Note that CSI constraints do not in principle require the specification of variable states (i.e., set of all states) even for variables that participate as contexts in such constraints. For example, the CSI (\(\mathsf{Light} \indep \mathsf{People}\ |\ \mathsf{Outage}=\textit{yes}\)) only requires ``yes" to be a valid state of the variable \(\mathsf{Outage}\) but does not require specifying the other states of \(\mathsf{Outage}\). Note also that earlier works 
in~\citep{nips/TikkaHK19,aistats/MokhtarianJEK22}, which studied the impact of CSI constraints on causal effect identifiability, 
did assume that all variables are binary which is effectively a state constraint. Proposition~\ref{prop:id-equiv} in Appendix~\ref{app:proof} shows that state constraints have no impact on variable-based or state-based identifiability without additional knowledge (e.g., CSIs), assuming the causal graph is semi-Markovian.\footnote{In a Semi-Markovian graph, every hidden variable is a root and has exactly two children~\citep{Verma93,uai/TianP02b}. This assumption is quite common in the existing literature, e.g., the \textsc{identify} algorithm in~\citep{TianPearl03,aaai/HuangV06} and the \textsc{ID} algorithm in~\citep{aaai/ShpitserP06}.} This perhaps explains why state constraints did not receive as much attention in the literature on causal effect identifiability.
We next consider the interplay between CSIs and state constraints though, showing that the latter can indeed improve both variable-based and state-based identifiability when combined with the former.

Consider the causal graph in Figure~\ref{sfig:state-eg1} which models flight delays. Here, \(\mathsf{Country}\ (U)\) refers to the country of departure and is assumed to be hidden. 
Assume further the CSI constraints (\(X \indep U\ |\ A \Equal \textit{short}\)) and (\(Y \indep U\ |\ X, A \Equal \textit{long}\)). Without restricting variable states, the variable-based causal effect \(\Pr_{X}(Y)\) is not identifiable.
Moreover, no state-based causal effect \(\Pr_{x}(y)\)
is identifiable in this case. 
However, if we assume the states of \(A\) to be \(\{\textit{short}, \textit{long}\}\), the causal effect \(\Pr_{X}(Y)\)
becomes identifiable and, hence, also state-based identifiable for any states \(x, y.\) In fact, we can estimate the causal effect using the following formula:
\(\Pr_{X}(Y)=\) \(\Pr(A \Equal \textit{short})\) \(\Pr(Y | X,A \Equal \textit{short})\) \(+\Pr(A \Equal \textit{long})\)\(\Pr(Y | X,A \Equal \textit{long}).\)
This example is formalized as Proposition~\ref{prop:cid-states} in Appendix~\ref{app:proof}. Interestingly, the state constraint of this example involves
an observed variable yet it still managed to improve identifiability when coupled with CSIs.

The next example shows that adding state constraints to CSIs may enable state-based identifiability but not variable-based identifiability. Consider the causal graph in Figure~\ref{sfig:state-eg2} which depicts a more detailed model of flight delays. Suppose \(U_1, U_2, U_3\) are hidden and we have the 
following CSIs: (\(A \indep\) \( U_1\ \)\(|\ X \Equal \textit{low-cost},\) \(U_2\)), (\(Y \indep\) \( U_3\ \)\(|\ B \Equal \textit{raining},\) \( U_2, A\)), and (\(Y \indep\) \( A\) \(|\ B \Equal \textit{snowing},\)\( U_2, U_3\)). Without state constraints, the causal effect \(\Pr_x(y)\) is unidentifiable for any states $x,y$ (hence, \(\Pr_X(Y)\) is not identifiable). If we assume that \(\mathsf{Weather}\ (B)\) has the only states \(\{\textit{raining}, \textit{snowing}\},\) then the causal effect becomes state-based identifiable but still not variable-based identifiable. In particular, the state constraint enables the identifiability of \(\Pr_{X \Equal \textit{low-cost}}(Y \Equal \textit{true})\) but not \(\Pr_{X \Equal \textit{high-cost}}(Y \Equal \textit{true})\) so \(\Pr_X(Y)\) remains unidentifiable. This is formalized as Proposition~\ref{prop:diff-domain} in Appendix~\ref{app:proof}.

\begin{figure}[tb]
\centering
\begin{subfigure}[b]{0.35\linewidth}
\centering
\begin{tikzpicture}[->=stealth,auto,scale=\dagsize,transform shape]
\node[rectangle,draw,font=\huge] (U) at (0,0) {Country ($U$)};
\node[font=\huge] (X) at (-2.5,-2) {Airline ($X$)};
\node[font=\huge] (B) at (2.5,-2) {Delay ($Y$)};
\node[font=\huge] (A) at (0,-4) {Distance ($A$)};

\path (U) edge (X);
\path (U) edge (B);
\path (X) edge (B);
\path (A) edge (X);
\path (A) edge (B);
\end{tikzpicture}
\caption{}
\label{sfig:state-eg1}
\end{subfigure}
\begin{subfigure}[b]{0.64\linewidth}
\centering
\begin{tikzpicture}[->=stealth,auto,scale=\dagsize,transform shape]
\node[rectangle,draw,font=\huge] (U1) at (-1.5,0) {Country ($U_1$)};
\node[font=\huge] (X) at (-2.5,-2.5) {Airline ($X$)};
\node[font=\huge] (A) at (2.5,-2.5) {Aircraft ($A$)};
\node[rectangle,draw,font=\huge] (U2) at (3,0) {Distance ($U_2$)};
\node[font=\huge] (Y) at (7.5,-2.5) {Delay ($Y$)};
\node[font=\huge] (B) at (7.5,0) {Weather ($B$)};
\node[rectangle,draw,font=\huge] (U3) at (3,-5) {Traffic ($U_3$)};

\path (U1) edge (X);
\path (U1) edge (A);
\path (U2) edge (A);
\path (U2) edge (Y);
\path (U3) edge (X);
\path (U3) edge (Y);
\path (X) edge (A);
\path (A) edge (Y);
\path (B) edge (Y);
\end{tikzpicture}
\caption{}
\label{sfig:state-eg2}
\end{subfigure}
\caption{causal graphs showing the effectiveness of state constraints when coupled with CSIs.}
\label{fig:state-eg}
\end{figure} 

\section{Testing Identifiability Under CSI and State Constraints}
\label{sec:main-algs}
We have so far discussed the role of additional constraints -- CSIs and state constraints -- in improving both variable-based and state-based identifiability. More importantly, we have demonstrated the separation between these two levels of identifiability, both under CSIs alone and when the two types of constraints are combined. We next introduce a systematic method for testing state-based identifiability in the presence of CSI and state constraints.  
We start by presenting an algorithm for testing identifiability under CSI constraints only, and then extend it in Section~\ref{ssec:state-alg} to treat state constraints as well. For the case of CSI constraints only, we immediately obtain a method for testing variable-based identifiability by considering specific states of treatments and outcomes based on Proposition~\ref{prop:state-var-eq}. For the case with additional state constraints, we can similarly test variable-based identifiability using the method for state-based identifiability as we demonstrate in Section~\ref{ssec:state-alg}. One assumption made by our method is that all contexts appearing in the CSIs are observed. This assumption commonly holds when the states of hidden variables are completely unknown (i.e., unconstrained) and therefore cannot appear in the CSI contexts.\footnote{This assumption is stronger than~\citep{nips/TikkaHK19} which allows hidden contexts, and is weaker than~\citep{aistats/MokhtarianJEK22} which assumes all contexts are observed roots in the graph.} 

Algorithm~\ref{alg:id-csi} shows the procedure for testing state-based identifiability, assuming the only constraints are CSIs. The main procedure \textsc{id-csi} takes as input the standard components for classical identifiability (a causal graph, observed variables, and treatment and outcome states), along with a set of CSI constraints.
The algorithm returns a formula for estimating the causal effect \(\Pr_\x(\y)\) if it is identifiable; otherwise, it returns \textsc{FAIL}. The structure of the algorithm is similar to that of \textsc{identify}~\citep{TianPearl03}: a causal graph is decomposed into subgraphs, called \emph{c-components}, and the problem of causal effect identification is converted to independent identification problems on these c-components. However, there are two key differences between Algorithm~\ref{alg:id-csi} and \textsc{identify}. First, Algorithm~\ref{alg:id-csi} computes the c-components of a novel notion called \emph{context-induced graph}, which is a pruned version of the causal graph based on CSIs (see Section~\ref{sec:i-ci-graph}). Second, the algorithm utilizes a modified procedure for the identification problem on c-components (of context-induced graphs) which is specified by the \textsc{id-qfunc} subroutine (see Section~\ref{sec:i-c-factor}). A detailed review of c-components and the associated notion of c-factors is provided in Appendix~\ref{app:ccomp}.

\begin{algorithm}[tb]
\footnotesize
\caption{State-Based Identifiability with CSI Constraints}
\label{alg:id-csi}
\begin{algorithmic}[1]
\Procedure{ID-CSI}{$G,$ $\V$, $\x$, $\y$, $\CC$}
\Statex // \textbf{input:} causal graph \(G,\) observed variables \(\V\), treatment \(\x\), outcome \(\y\), CSI constraints \(\CC\)
\Statex // \textbf{output:} identifying formula for \(\Pr_\x(\y)\) or FAIL
\Statex // Assume \(\V = (V^1, V^2, \dots)\) are ordered topologically in \(G\), and let \(\V^{(i)} = \{V^1, \dots, V^i\}\)
\ForEach{CI-graph \(T\) under context \(\x\y\)}
\State \(M \gets \) remove \(\X\) and retain the subgraph of $T$ consisting of all ancestors of \(\Y\)
\State \(\C_1, \dots \C_m \gets\) maximal c-components in \(M\)
\For{$i=1$ to \(m\)}
\ForEach{context \(\s\) and its CI-graph \(H\)}
\If{$\C_i$ have the same parents in \(T\) and \(H\)
 \textbf{and} $(\s \setminus \x\y)$ contains no state of \(\C_i\) or their parents}
\State \(f^i \gets \Call{ID-Qfunc}{}\)\((\C_i, \V \cup \U, \Pr(\V),\) \(\s, H)\) \;\;\;\; // \(\U\) are the hidden variables
\If{\(f^i \neq \text{FAIL}\)} \textbf{break} \EndIf
\EndIf
\EndFor
\EndFor
\If{\(f^i \neq \text{FAIL}\) for $i=1, \dots, m$}
\State \Return \(\sum_{\V \setminus (\X\cup\Y)}\prod_{i=1}^m f^i_{\x\y}\) 
\EndIf
\EndFor
\State \Return FAIL
\EndProcedure
\Statex
\Procedure{ID-Qfunc}{$\C$, $\D$, $f$, $\s$, $H$}
\Statex // \textbf{input:} c-component \(\C\), variable set \(\D\), expression (factor) $f$, context \(\s\), graph \(H\)
\Statex // \textbf{output:} formula for computing the c-factor \(f^i\) or FAIL
\If{\(\D = \C\)} \Return \(f_\s\)
\EndIf
\State\(H_\D \gets\) subgraph of \(H\) formed by variables in \(\D\)
\If{exists a leaf node \(W\) in \(H_\D\) and \(W \notin \S \cup \C\)}
\State \(\D' \gets \D \setminus \{W\}\)
\State \(f' \gets \sum_W f\)
\State \Return \Call{ID-Qfunc}{$\C, \D', f', \s, H$}
\EndIf
\If{exists a maximal c-component \(\D'\) in \(H_\D\) such that \(\C \subseteq \D' \subset \D\)}
\State \(f' \gets \prod_{V^{i} \in (\D' \cap \V)}\frac{\sum_{\D \setminus \V^{(i)}}f}{\sum_{\D \setminus \V^{(i-1)}}f}\) 
\State \Return \Call{ID-Qfunc}{$\C, \D', f', \s, H$}
\EndIf
\State \Return FAIL
\EndProcedure
\end{algorithmic}
\end{algorithm}

\subsection{Identifying Causal Effects Using Context-Induced Graphs}
\label{sec:i-ci-graph}

We start with the notion of context-induced graphs (utilized by Algorithm~\ref{alg:id-csi}), which simplify the causal graph by leveraging CSI constraints. The definition of context-induced graph (CI-graph) takes as input a set of CSIs and a context and removes edges based on the CSIs activated by the context.

\begin{definition}
\label{def:ci-graph}
Let \(G\) be a causal graph, \(\CC\) be a set of CSIs, and \(\s\) be a context. A \underline{CI-graph} under \(\s\) is obtained from \(G\) by repeatedly removing edges as follows: remove edge \(P_1 \rightarrow C\) if there exists a CSI \((C \indep P_1\ |\ \p_2, \P_3)\) for which \(\p_2 \subseteq \s\) and \(\P_2\) are parents of \(C\) at the time of removal.\footnote{The notion of CI-graph differs from existing notions that leverage CSIs; see footnote~\ref{ft:ci-graph} in the appendix for more details.}
\end{definition}

Given a fixed context, the choice of CI-graph may not be unique. To illustrate, consider the causal graph in Figure~\ref{sfig:prune1} and assume the CSIs \((D \indep B\ |\ c_0, A)\), \((D \indep C\ |\ b_0, A),\) and \((E \indep B\ |\ c_0).\) The graphs in Figures~\ref{sfig:prune2} and~\ref{sfig:prune3} are both valid CI-graphs under the context \(b_0c_0.\) In particular, Figures~\ref{sfig:prune2} is obtained by first removing the edge \(B \rightarrow E\) based on the CSI \((E \indep B\ |\ c_0),\) and then removing the edge \(B \rightarrow D\) based on \((D \indep B\ |\ c_0, A)\). On the other hand, Figures~\ref{sfig:prune2} applies the same first step but removes the edge \(C \rightarrow D\) based on \((D \indep C\ |\ b_0, A)\) as the second step. No more edges can be removed from either graph. CI-graphs 
reduce the parameters of a causal graph by removing some of its edges, while 
preserving the (conditional) independencies implied by the original causal graph and CSI constraints given the context; see Section~\ref{sec:ci-graph} for further details.

\begin{figure}[tb]
\centering
\begin{subfigure}[b]{0.24\linewidth}
\centering
\begin{tikzpicture}[->=stealth,auto,scale=\dagsize,transform shape]
\node[state,font=\huge] (A) at (0,0) {$A$};
\node[font=\huge] (C) at (-1,-2) {$Y$};
\node[state,font=\huge] (B) at (2,0) {$D$};
\node[font=\huge] (D) at (1,-2) {$J$};
\node[font=\huge] (E) at (3,-2) {$S$};
\path (A) edge (C);
\path (A) edge (D);
\path (B) edge (D);
\path (D) edge (E);
\path (C) edge[bend right=20] (E);
\end{tikzpicture}
\caption{CI-graph~\ref{sfig:csi-eg1}}
\label{sfig:ci-mut-eg12}
\end{subfigure}
\begin{subfigure}[b]{0.24\linewidth}
\centering
\begin{tikzpicture}[->=stealth,auto,scale=\dagsize,transform shape]
\node[state,font=\huge,red] (A) at (0,0) {$A$};
\node[state,font=\huge,red] (B) at (2,0) {$D$};
\node[font=\huge,red] (D) at (1,-2) {$J$};
\node[font=\huge,blue] (E) at (3,-2) {$S$};
\path (A) edge (D);
\path (B) edge (D);
\path (D) edge (E);
\end{tikzpicture}
\caption{mutilated~\ref{sfig:csi-eg1}}
\label{sfig:ci-mut-eg13}
\end{subfigure}
\begin{subfigure}[b]{0.24\linewidth}
\centering
\begin{tikzpicture}[->=stealth,auto,scale=\dagsize,transform shape]
\node[state,font=\huge] (A) at (0,0) {$A$};
\node[font=\huge] (C) at (-1,-2) {$Y$};
\node[state,font=\huge] (B) at (2,0) {$D$};
\node[font=\huge] (D) at (1,-2) {$J$};
\node[font=\huge] (E) at (3,-2) {$S$};
\path (A) edge (C);
\path (A) edge (D);
\path (B) edge (E);
\path (D) edge (E);
\path (C) edge (D);
\end{tikzpicture}
\caption{CI-graph~\ref{sfig:csi-eg2}}
\label{sfig:ci-mut-eg22}
\end{subfigure}
\begin{subfigure}[b]{0.24\linewidth}
\centering
\begin{tikzpicture}[->=stealth,auto,scale=\dagsize,transform shape]
\node[state,font=\huge,red] (B) at (3,0) {$D$};
\node[font=\huge,red] (E) at (4.5,-2) {$S$};
\path (B) edge (E);
\end{tikzpicture}
\caption{mutilated~\ref{sfig:csi-eg2}}
\label{sfig:ci-mut-eg23}
\end{subfigure}
\caption{context-induced and mutilated graphs for Figure~\ref{fig:csi-eg}.}
\label{fig:ci-mut-eg}
\end{figure} 

Instead of testing identifiability using the original causal graph, Algorithm~\ref{alg:id-csi} enumerates all possible CI-graphs \(T\) under the context \(\x\y\) and checks whether the causal effect can be identified using any of them (line~2). Similar to the classical identification method \textsc{identify}~\citep{TianPearl03}, the CI-graph \(T\) is then mutilated by removing the treatments \(\X\) and retaining the subgraph consisting of the ancestors of outcomes \(\Y\) (line~3). If we decompose the resulting graph into c-components \(\{\C_1, \dots \C_m\}\), we obtain the following key property: \(\Pr_\x(\y) = \sum_{\V \setminus (\X \cup \Y)} \prod_{i=1}^m f^i_{\x\y}\), where \(f^i_{\x\y}\) denotes the entries of the c-factor for \(\C_i\) that are compatible with \(\x\y.\)\footnote{The c-factor for each c-component \(\C_i\) in graph \(T\) is defined as \(f^i = \sum_{\U_i}\prod_{C \in \C_i} \theta_{C | \P_C},\) where \(\U_i\) denotes the hidden variables in \(\C_i\) and \(\P_C\) denotes the parents of \(C\) in graph \(T.\) For example, in Figure~\ref{sfig:ci-mut-eg12}, the c-factor for the c-component \(\{A,D,J\}\) can be computed as \(f = \sum_{AD}\theta_A\theta_D\theta_{J|AD}\), and its entries under the context \(y_0j_0s_0\) are given by \(f_{y_0j_0s_0} = \sum_{AD}\theta_A\theta_D\theta_{j_0|AD}.\) See Appendix~\ref{app:ccomp} for additional examples.} Hence, if all c-factors corresponding to the c-components in the (mutilated) CI-graph are identifiable from the observational distribution \(\Pr(\V)\), the causal effect is also identifiable. This yields a sound method for testing identifiability by checking identifiability with respect to each independent c-component in the CI-graph. We state this result formally as Lemma~\ref{thm:id-ccomp} with proof in Appendix~\ref{app:proof}. 

To illustrate, consider the causal graph in Figure~\ref{sfig:csi-eg1} with the CSI \((S \indep D\ |\ \) \(Y, j_0)\) and state-based causal effect \(\Pr_{y_0}(j_0, s_0)\). Figures~\ref{sfig:ci-mut-eg12} and~\ref{sfig:ci-mut-eg13} depict a CI-graph under the context \(y_0j_0s_0\) (treatment and outcome states) and its mutilated graph, where \(\{A, D, J\}\) and \(\{S\}\) are the c-components in the mutilated graph (highlighted with colors). Based on our discussion earlier, the causal effect is identifiable if the corresponding c-factors are identifiable. Consider now a different example in Figure~\ref{sfig:csi-eg2} with CSI \((S \indep D\ |\ Y, j_0)\) and causal effect \(\Pr_{y_0}(j_0,s_0).\) Figures~\ref{sfig:ci-mut-eg22} and~\ref{sfig:ci-mut-eg23} depict the CI-graph and its mutilated graph under the context \(y_0j_0s_0.\) Since \(\{D,S\}\) is the only c-component in the mutilated graph, the causal effect is identifiable if the c-factor is identifiable. 
One key observation is that, since CI-graphs are always subgraphs of the original graph, the c-components (and corresponding c-factors) extracted from the mutilated CI-graphs are no larger than those in the original graph, which intuitively allows more causal effects to be identified.

\subsection{Treating C-Components of CI-Graphs}
\label{sec:i-c-factor}

We have shown that the problem of identifying causal effects can be converted to that of identifying c-factors induced by the c-components in the CI-graph. We next present a method for testing the identifiability of these c-factors, which involves the \textsc{id-qfunc} procedure in Algorithm~\ref{alg:id-csi}.

\begin{figure}[tb]
\centering
\begin{subfigure}[b]{0.24\linewidth}
\centering
\begin{tikzpicture}[->=stealth,auto,scale=\dagsize,transform shape]
\node[state,font=\huge] (U) at (0,0) {$A$};
\node[font=\huge] (Y) at (1,-2) {$D$};
\node[font=\huge] (X1) at (-1,-2) {$B$};
\node[font=\huge] (X2) at (2,0) {$C$};
\node[font=\huge] (E) at (3,-2) {$E$};

\path (U) edge (Y);
\path (U) edge (X1);
\path (X1) edge (Y);
\path (X2) edge (Y);
\path (X2) edge (E);
\path (X1) edge[bend right=20] (E);
\end{tikzpicture}
\caption{original $G$}
\label{sfig:prune1}
\end{subfigure}
\begin{subfigure}[b]{0.24\linewidth}
\centering
\begin{tikzpicture}[->=stealth,auto,scale=\dagsize,transform shape]
\node[state,font=\huge] (U) at (0,0) {$A$};
\node[font=\huge] (Y) at (1,-2) {$D$};
\node[font=\huge] (X1) at (-1,-2) {$B$};
\node[font=\huge] (X2) at (2,0) {$C$};
\node[font=\huge] (E) at (3,-2) {$E$};

\path (U) edge (Y);
\path (U) edge (X1);
\path (X2) edge (Y);
\path (X2) edge (E);
\end{tikzpicture}
\caption{CI-graph \(T\)}
\label{sfig:prune2}
\end{subfigure}
\begin{subfigure}[b]{0.24\linewidth}
\centering
\begin{tikzpicture}[->=stealth,auto,scale=\dagsize,transform shape]
\node[state,font=\huge] (U) at (0,0) {$A$};
\node[font=\huge] (Y) at (1,-2) {$D$};
\node[font=\huge] (X1) at (-1,-2) {$B$};
\node[font=\huge] (X2) at (2,0) {$C$};
\node[font=\huge] (E) at (3,-2) {$E$};

\path (U) edge (Y);
\path (U) edge (X1);
\path (X1) edge (Y);
\path (X2) edge (E);
\end{tikzpicture}
\caption{CI-graph \(H\)}
\label{sfig:prune3}
\end{subfigure}
\begin{subfigure}[b]{0.24\linewidth}
\centering
\begin{tikzpicture}[->=stealth,auto,scale=\dagsize,transform shape]
\node[state,font=\huge,red] (U) at (0,0) {$A$};
\node[font=\huge,red] (Y) at (1,-2) {$D$};
\node[font=\huge,blue] (X2) at (2,0) {$C$};

\path (U) edge (Y);
\path (X2) edge (Y);
\end{tikzpicture}
\caption{mutilated \(T\)}
\label{sfig:prune4}
\end{subfigure}
\caption{causal graph, distinct context-induced graphs, and a mutilated graph under the context \(b_0c_0.\)}
\label{fig:prune}
\end{figure}

A graph-based method for testing c-factor identifiability was introduced in \textsc{identify}~\citep{TianPearl03}. This method applies to the c-factors obtained from the original causal graphs but does not extend to c-factors corresponding to the c-components of CI-graphs. We illustrate this limitation with a concrete example in Appendix~\ref{app:proof} Proposition~\ref{prop:incorrect-ci-graph}. Here, we propose a novel method for treating the identification of c-components (and their corresponding c-factors) within CI-graphs by modifying the approach in \textsc{identify}. As we demonstrate later, our method is sound for c-factor identification, and is complete for classical identifiability (without constraints) when combined with the procedure described in Section~\ref{sec:i-ci-graph}.

Our goal is to check the identifiability of the c-factor for each c-component \(\C_i\) in the (mutilated) CI-graph \(T.\) To achieve this, we start by enumerating all possible contexts \(\s\) and their CI-graphs \(H\) (line~6). We then check whether (i) variables in \(\C_i\) have the same parents in \(T\) and \(H\), and (ii) \((\s \setminus \x\y)\) contains no state of \(\C_i\) or of their parents (line~7). The intuition here is that if both conditions are satisfied, the c-factor induced by \(\C_i\) in \(H\) (denoted \(f^i_{\s}\)) must contain all the entries of the c-factor of interest (denoted \(f^i_{\x\y}\)).\footnote{The two c-factors contain the same set of CPTs by condition~(i). Moreover, if \(\s\) contains the state of some variable in the CPTs, that state must appear in \(\x\y\) by condition~(ii), which implies that the entries in \(f^i_{\s}\) form a superset of those in \(f^i_{\x\y}.\)} Hence, we can instead check the identifiability of \(f^i_{\s}\) by invoking the \textsc{id-qfunc} subroutine (line~8).
For example, suppose Figures~\ref{sfig:prune2} and~\ref{sfig:prune3} depict graphs \(T\) and \(H\) induced by contexts \(c_0e_0\) and \(b_0c_0\). Both conditions hold for the c-component \(\{E\}\) since \(E\) has the same parent \(C\) in the graphs and \((b_0c_0 \setminus c_0e_0)\) does not involve \(E\) or its parent. In contrast, condition (i) is violated for c-component \(\{A,B,D\}\).

The \textsc{id-qfunc} checks the identifiability of c-factor \(f^i_{\s}\) recursively. It initializes a variable set \(\D\) containing all variables ($\V \cup \U$) and iteratively modifies it using the following two operations. Let \(H_\D\) be the subgraph of \(H\) containing variables in \(\D\) and all edge between them. At each step, either (1) remove a leaf node in \(H_\D\) that is not in \(\S \cup \C_i\) (lines~16-19); or (2) replace \(\D\) by a maximal c-component \(\D'\) in \(H_\D\) such that \(\D\supset \D' \supseteq \C_i\) (lines~20-22).  If the recursion terminates with \(\D = \C_i\), the c-factor is identifiable, and \textsc{id-qfunc} returns a formula for its estimation; otherwise, it returns FAIL. We formulate this c-component identification method as Lemma~\ref{thm:cc-id} in Appendix~\ref{app:proof}.

We now illustrate how this identifiability method can be applied to treat the c-components in the CI-graph (Section~\ref{sec:i-ci-graph}). Consider again the c-components \(\{A,D,J\}\) and \(\{S\}\) in the (mutilated) CI-graph in Figure~\ref{sfig:ci-mut-eg13} under the context \(y_0j_0s_0.\) To identify these c-components, let \(j_0\) be the context \(\s\) in line~6 and Figure~\ref{sfig:ci-mut-eg12} be its CI-graph \(H\). Since \(\s\) and \(H\) satisfy the conditions in line~7, the causal effect is identifiable if \textsc{id-qfunc} establishes identifiability for both c-components. This is indeed the case. The first c-component \(\{A,D,J\}\) can be obtained by replacing \((\V \cup \U)\) with the maximal c-component \(\{A,D,J,Y\}\) and then pruning the leaf node \(Y\). The second c-component \(\{S\}\) is already a maximal c-component of \(T\).

We now formally state the soundness of Algorithm~\ref{alg:id-csi}.
\begin{theorem}
\label{thm:sound}
Let \(\CC\) be a set of CSI constraints. Then \(\Pr_\x(\y)\) is identifiable wrt \(\langle G, \V, \CC \rangle\) if Algorithm~\ref{alg:id-csi} returns a formula for estimating \(\Pr_\x(\y)\).
\end{theorem}

\begin{algorithm}[tb]
\footnotesize
\caption{State-Based Identifiability with CSI and State Constraints}
\label{alg:id-csi-state}
\begin{algorithmic}[1]
\Procedure{ID-CSI-State}{$G,$ $\V$, $\x$, $\y$, $\CC_1$, $\CC_2$}
\Statex // \textbf{input:} graph \(G,\) observed variables \(\V\), treatment \(\x\), outcome \(\y\), CSI constraints \(\CC_1\), state constraints \(\CC_2\)
\Statex // \textbf{output:} identifying formula for \(\Pr_\x(\y)\) or FAIL
\For{each possible context-partition \(\{\s_1, \dots, \s_k\}\) of \(\V \setminus (\X \cup \Y)\) according to \(\CC_2\)}
\For{$i=1$ to \(k\)}
\State \(\Pr_{\x}(\y, \s_i) \gets \Call{ID-CSI}{G, \CC_1, \x, \y, \V}\)
\EndFor
\If{$\Pr_{\x}(\y, \s_i) \neq \text{FAIL}$ for $i=1, \dots, k$}
\State \Return $\sum_{i=1}^k \Pr_{\x}(\y, \s_i)$
\EndIf
\EndFor
\State \Return FAIL
\EndProcedure
\end{algorithmic}
\end{algorithm}

If the set of constraints $\CC$ is empty, then Algorithm~\ref{alg:id-csi} is both sound and complete for testing identifiability. That is, when $\CC$ is empty, Algorithm~\ref{alg:id-csi}  can identify all causal effects that are identifiable by classical identification methods.

\begin{proposition}
\label{prop:id-complete}
\(\Pr_\x(\y)\) is identifiable wrt \(\langle G, \V \rangle\) iff it is identifiable by Algorithm~\ref{alg:id-csi}.
\end{proposition}

In general, for a causal graph with \(n\) variables, the time complexity of Algorithm~\ref{alg:id-csi} can be exponential in $n$, since the numbers of possible CI-graphs (line~2) and contexts (line~6) can be exponential in the worst case. If the total number of contexts appearing in CSIs is bounded by a constant, the worst-case complexity reduces to \(O(n^3).\)\footnote{It suffices to enumerate \(\s\) for which all states appear in the contexts of CSIs in line~6 without compromising the completeness. The complexity is dominated by \textsc{id-qfunc} in this case. In general, testing causal effect identifiability under CSIs with binary variables is NP-hard~\citep{nips/TikkaHK19}.}

We have introduced a sound algorithm for testing state-based (variable-based) identifiability \emph{when the only constraints are CSIs.} When the causal effect is identifiable, the algorithm also returns a formula for estimating the causal effect. We next discuss an extension of this algorithm that further enables leveraging state constraints in conjunction with CSIs.

\subsection{Leveraging State Constraints}
\label{ssec:state-alg}

\begin{figure}[tb]
\centering
\begin{subfigure}[b]{0.24\linewidth}
\centering
\begin{tikzpicture}[->=stealth,auto,scale=\dagsize,transform shape]
\node[state,font=\huge] (A) at (0,0) {$U$};
\node[font=\huge] (C) at (-1.5,-1.5) {$X$};
\node[font=\huge] (D) at (1.5,-1.5) {$Y$};
\node[font=\huge] (B) at (0,-3) {$A$};

\path (A) edge (D);
\path (B) edge (C);
\path (B) edge (D);
\path (C) edge (D);
\end{tikzpicture}
\caption{context \(a_0\)}
\label{sfig:state1-2}
\end{subfigure}
\begin{subfigure}[b]{0.24\linewidth}
\centering
\begin{tikzpicture}[->=stealth,auto,scale=\dagsize,transform shape]
\node[state,font=\huge] (A) at (0,0) {$U$};
\node[font=\huge] (C) at (-1.5,-1.5) {$X$};
\node[font=\huge] (D) at (1.5,-1.5) {$Y$};
\node[font=\huge] (B) at (0,-3) {$A$};

\path (A) edge (C);
\path (B) edge (C);
\path (B) edge (D);
\path (C) edge (D);
\end{tikzpicture}
\caption{context \(a_1\)}
\label{sfig:state1-3}
\end{subfigure}
\begin{subfigure}[b]{0.24\linewidth}
\centering
\begin{tikzpicture}[->=stealth,auto,scale=\dagsize,transform shape]
\node[state,font=\huge] (A) at (0,0) {$U_1$};
\node[state,font=\huge] (B) at (2,0) {$U_2$};
\node[font=\huge] (D) at (4,0) {$B$};
\node[font=\huge] (E) at (0,-2) {$X$};
\node[font=\huge] (F) at (2,-2) {$A$};
\node[font=\huge] (G) at (4,-2) {$Y$};
\node[state,font=\huge] (C) at (2,-4) {$U_3$};

\path (A) edge (E);
\path (B) edge (F);
\path (B) edge (G);
\path (D) edge (G);
\path (E) edge (F);
\path (F) edge (G);
\path (C) edge (E);
\end{tikzpicture}
\caption{context \(b_0\)}
\label{sfig:state2-2}
\end{subfigure}
\begin{subfigure}[b]{0.24\linewidth}
\centering
\begin{tikzpicture}[->=stealth,auto,scale=\dagsize,transform shape]
\node[state,font=\huge] (A) at (0,0) {$U_1$};
\node[state,font=\huge] (B) at (2,0) {$U_2$};
\node[font=\huge] (D) at (4,0) {$B$};
\node[font=\huge] (E) at (0,-2) {$X$};
\node[font=\huge] (F) at (2,-2) {$A$};
\node[font=\huge] (G) at (4,-2) {$Y$};
\node[state,font=\huge] (C) at (2,-4) {$U_3$};

\path (A) edge (E);
\path (B) edge (F);
\path (B) edge (G);
\path (D) edge (G);
\path (E) edge (F);
\path (C) edge (E);
\path (C) edge (G);
\end{tikzpicture}
\caption{context \(b_1\)}
\label{sfig:state2-3}
\end{subfigure}

\begin{subfigure}[b]{0.49\linewidth}
\centering
\begin{tikzpicture}[->=stealth,auto,scale=\dagsize,transform shape]
\node[font=\huge] (A) at (-1,0) {$A$};
\node[font=\huge] (B) at (3,0) {$B$};
\node[state,font=\huge] (C) at (-2,-1.5) {$C$};
\node[font=\huge] (E) at (-3,-3) {$E$};
\node[state,font=\huge] (F) at (1,-3) {$F$};
\node[font=\huge] (G) at (5,-3) {$G$};
\node[state,font=\huge] (D) at (1,-1.5) {$D$};

\path (A) edge (C);
\path (C) edge (E);
\path (A) edge (F);
\path (C) edge (F);
\path (E) edge (F);
\path (B) edge (F);
\path (B) edge (G);
\path (D) edge (E);
\path (D) edge (G);
\path (F) edge (G);
\end{tikzpicture}
\caption{original}
\label{sfig:state3-1}
\end{subfigure}
\begin{subfigure}[b]{0.49\linewidth}
\centering
\begin{tikzpicture}[->=stealth,auto,scale=\dagsize,transform shape]
\node[font=\huge] (A) at (-1,0) {$A$};
\node[font=\huge] (B) at (3,0) {$B$};
\node[state,font=\huge] (C) at (-2,-1.5) {$C$};
\node[font=\huge] (E) at (-3,-3) {$E$};
\node[state,font=\huge] (F) at (1,-3) {$F$};
\node[font=\huge] (G) at (5,-3) {$G$};
\node[state,font=\huge] (D) at (1,-1.5) {$D$};

\path (A) edge (C);
\path (C) edge (E);
\path (A) edge (F);
\path (C) edge (F);
\path (E) edge (F);
\path (B) edge (F);
\path (B) edge (G);
\path (D) edge (E);
\path (D) edge (G);
\end{tikzpicture}
\caption{context \(b_0\)}
\label{sfig:state3-2}
\end{subfigure}
\begin{subfigure}[b]{0.49\linewidth}
\centering
\begin{tikzpicture}[->=stealth,auto,scale=\dagsize,transform shape]
\node[font=\huge] (A) at (-1,0) {$A$};
\node[font=\huge] (B) at (3,0) {$B$};
\node[state,font=\huge] (C) at (-2,-1.5) {$C$};
\node[font=\huge] (E) at (-3,-3) {$E$};
\node[state,font=\huge] (F) at (1,-3) {$F$};
\node[font=\huge] (G) at (5,-3) {$G$};
\node[state,font=\huge] (D) at (1,-1.5) {$D$};

\path (A) edge (C);
\path (C) edge (E);
\path (A) edge (F);
\path (E) edge (F);
\path (B) edge (F);
\path (B) edge (G);
\path (D) edge (E);
\path (F) edge (G);
\end{tikzpicture}
\caption{context \(a_0b_1\)}
\label{sfig:state3-3}
\end{subfigure}
\begin{subfigure}[b]{0.49\linewidth}
\centering
\begin{tikzpicture}[->=stealth,auto,scale=\dagsize,transform shape]
\node[font=\huge] (A) at (-1,0) {$A$};
\node[font=\huge] (B) at (3,0) {$B$};
\node[state,font=\huge] (C) at (-2,-1.5) {$C$};
\node[font=\huge] (E) at (-3,-3) {$E$};
\node[state,font=\huge] (F) at (1,-3) {$F$};
\node[font=\huge] (G) at (5,-3) {$G$};
\node[state,font=\huge] (D) at (1,-1.5) {$D$};

\path (A) edge (C);
\path (C) edge (E);
\path (A) edge (F);
\path (C) edge (F);
\path (B) edge (F);
\path (B) edge (G);
\path (D) edge (E);
\path (F) edge (G);
\end{tikzpicture}
\caption{context \(a_1b_1\)}
\label{sfig:state3-4}
\end{subfigure}
\caption{CI-graphs under different contexts in the context-partition for various examples.}
\label{fig:state-alg1}
\end{figure}

As demonstrated in Section~\ref{sec:domain-constraints}, additional causal effects become identifiable when CSIs are coupled with state constraints. We next propose a systematic approach for testing state-based (and variable-based) identifiability under both CSI and state constraints by leveraging Algorithm~\ref{alg:id-csi}. Note that we allow variables with arbitrary cardinalities, which is more general than~\citep{nips/TikkaHK19} and~\citep{aistats/MokhtarianJEK22} which assume all variables are binary.

The detailed procedure is shown in Algorithm~\ref{alg:id-csi-state} which relies on an important notion called \emph{context-partitions}. In particular, we say that a set \(\{\s_1, \dots \s_k\}\) forms a context-partition of variables \(\V\) if (i) each $\s_i,$ called a context, is an instantiation of some variables in $\V,$
(ii) $\s_i$ and $\s_j$ are not compatible for $i \neq j$, and 
(iii) every instantiation $\v$ of $\V$ is compatible with some context $\s_i.$
That is, contexts do not overlap and cover all instantiations of \(\V.\) To illustrate, suppose variables \(A, B, C\) have states \(\{a_0,a_1,a_2\}\), \(\{b_0,b_1\}\) and \(\{c_0,c_1\}\), respectively. Then \(\{a_0,\:a_1b_0,\:a_1b_1,\: a_2c_1,\:a_2c_2\}\) 
is a context-partition of these variables and so is 
\(\{a_0b_0,\:a_0b_1,\:a_1,\:a_2b_0c_0,\:a_2b_0c_1,\:a_2b_1\}.\)

Algorithm~\ref{alg:id-csi-state} enumerates all possible context-partitions \(\{\s_1, \dots, \s_k\}\) over the variable states specified by state constraints, and reduces the problem of identifying \(\Pr_\x(\y)\) to that of identifying sub-queries \(\Pr_\x(\y, \s_i)\) for each context \(\s_i\) (line~2). It then invokes Algorithm~\ref{alg:id-csi} as a subroutine to check the identifiability of these sub-queries (lines 3-6). If all sub-queries are deemed identifiable, \(\Pr_\x(\y)\) is also identifiable. The intuition is that, by augmenting the treatment and outcome states \(\x\y\) with additional contexts \(\s_i\), we can exploit additional CSIs whose contexts extend beyond \(\x\y\) and hence identify more causal effects. When the causal effect is identifiable, the algorithm also returns a formula for estimating it using the equality \(\Pr_\x(\y) = \sum_{i} \Pr(\y, \s_i).\)

\begin{figure}[tb]
\centering
\includegraphics[width=1\linewidth]{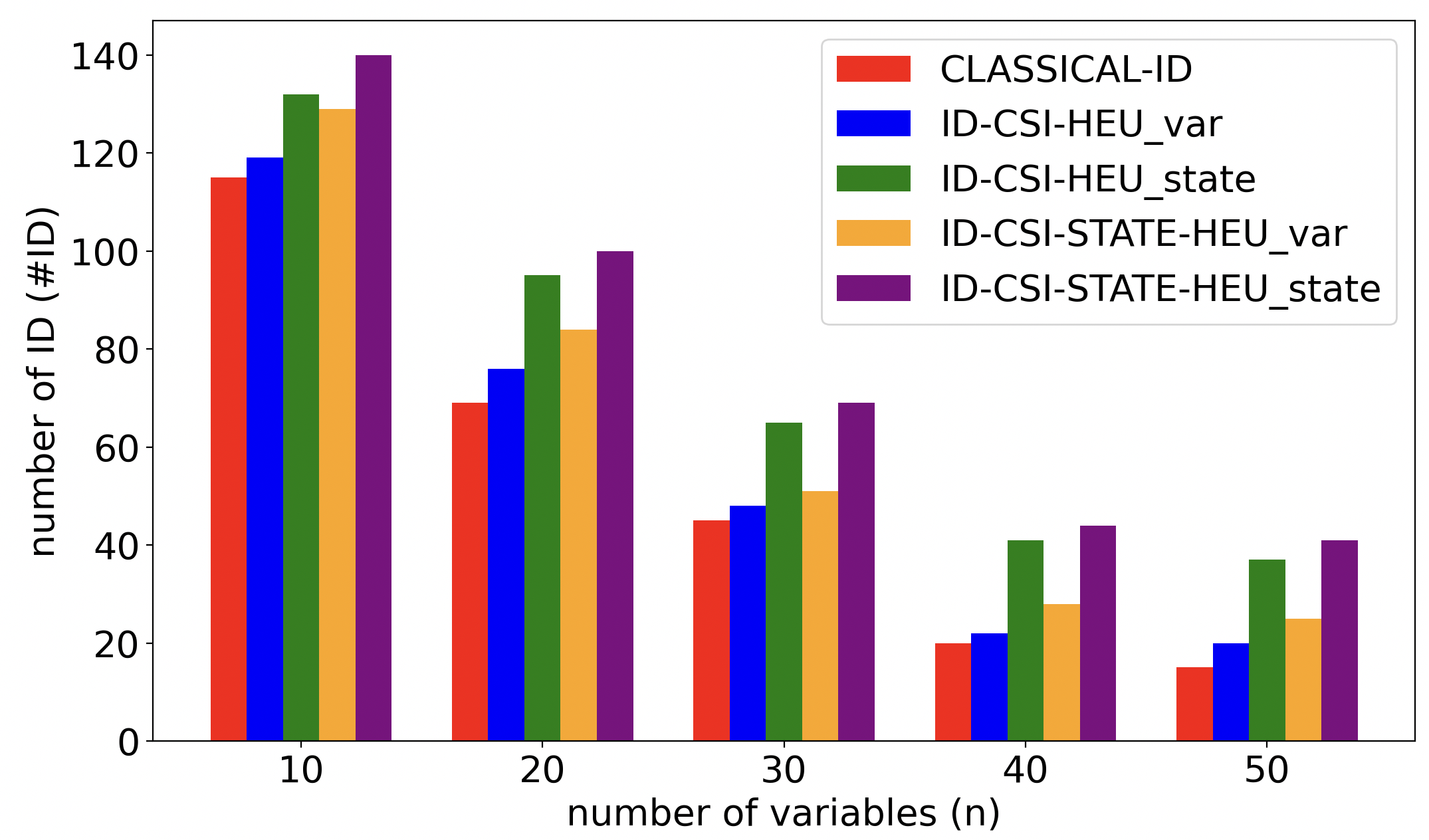}
\caption{number of identifiable causal effects (out of 200 runs) as the number of variables increases with $r=0.8.$}
\label{fig:exp}
\end{figure}

We now state the soundness of Algorithm~\ref{alg:id-csi-state} for identifiability under CSI and state constraints.

\begin{theorem}
\label{thm:partition}
Let \(\CC_1\) be a set of CSIs and \(\CC_2\) a set of state constraints. Then \(\Pr_\x(\y)\) is identifiable wrt \(\langle G, \V, \CC_1 \cup \CC_2 \rangle\) if Algorithm~\ref{alg:id-csi-state} returns a formula for estimating \(\Pr_\x(\y).\)
\end{theorem}

We next demonstrate how Algorithm~\ref{alg:id-csi-state} can be applied to identify causal effects when state constraints are available. First consider the causal graph in Figure~\ref{sfig:state-eg1} with the CSI constraints \((X \indep U\ |\ a_0)\), \((Y \indep U\ |\ X, a_1)\). Suppose variable \(A\) only has two states \(\{a_0,a_1\}.\) Instead of directly checking the identifiability of the original \(\Pr_{x_0}(y_0)\), we split the causal effect into sub-queries \(\Pr_{x_0}(y_0,a_0)\) and \(\Pr_{x_0}(y_0,a_1)\) according to the context-partition \(\{a_0, a_1\}.\) Figures~\ref{sfig:state1-2} and~\ref{sfig:state1-3} depict the CI-graphs for the sub-queries. Since both sub-queries are identifiable by Algorithm~\ref{alg:id-csi}, the original causal effect is also identifiable; see the proof of Proposition~\ref{prop:cid-states} for more details. 
Similarly, we can apply the theorem to identify \(\Pr_{x_0}(y_0)\) in Figure~\ref{sfig:state-eg2} by constructing two sub-queries whose CI-graphs are shown in~\Cref{sfig:state2-2,sfig:state2-3}; see the proof of Proposition~\ref{prop:diff-domain} for details. 

We now show another example with the causal graph in Figure~\ref{sfig:state3-1} that requires splitting more than one variable in the context-partition. Suppose the states for variables \(A, B\) are \(\{a_0, a_1\}\), \(\{b_0, b_1\}\) and assume the following CSI constraints: \((G \indep F\ |\ b_0, D)\), \((G \indep D\ |\ b_1, F),\) \((F \indep C\ |\ a_0, B, E),\) \((F \indep E\ |\ a_1, B, C).\) The state-based causal effect \(\Pr_{e_0}(g_0)\) (and variable-based \(\Pr_E(G)\)) is identifiable since each sub-query induced by the context-partition \(\{b_0, a_0b_1, a_1b_1\}\) can be identified using Algorithm~\ref{alg:id-csi}; see~\Cref{sfig:state3-2,sfig:state3-3,sfig:state3-4} for the CI-graphs under each context in the partition.

We next discuss a method for testing \emph{variable-based} identifiability under both CSI and state constraints. Similar to the case with CSI constraints only (Proposition~\ref{prop:state-var-eq} in Section~\ref{sec:csi}), the method is based on reducing variable-based identifiability to state-based identifiability. Specifically, to check the identifiability of \(\Pr_\X(\Y)\), we consider instantiations \(\x, \y\) in which each state is either included in the state constraints or outside the CSI contexts. The number of such instantiations is finite if every variable outside the state constraints is assigned the same state (outside the CSI contexts) across all instantiations. Then \(\Pr_\X(\Y)\) is identifiable iff \(\Pr_\x(\y)\) is identifiable for all these instantiations. We formalize this result as Proposition~\ref{prop:state-var-eq-ext} in Appendix~\ref{app:proof}.

\subsection{Heuristics}
\label{ssec:heuristics}
Enumerating CI-graphs (Algorithm~\ref{alg:id-csi}, line~2), contexts and their corresponding CI-graphs (Algorithm~\ref{alg:id-csi}, lines~6-7), and context-partitions (Algorithm~\ref{alg:id-csi-state} line~2) incur exponential time complexity and are therefore impractical. To address this issue, we propose in Appendix~\ref{app:heuristics}
heuristics to select a single choice for each step, avoiding the need to enumerate all possible choices. The heuristics reduce the complexity of Algorithm~\ref{alg:id-csi} from exponential to polynomial time. Algorithm~\ref{alg:id-csi-state}, however, can still require exponential time, since the number of contexts in the context-partition can be exponential in the worst case. \Cref{alg:id-csi,alg:id-csi-state} combined with these heuristics will be denoted by \textsc{id-csi-heu} and \textsc{id-csi-state-heu}. We will use these versions of the algorithms in our upcoming experiments 
which demonstrate that the heuristics substantially improve the computational efficiency of our proposed identification algorithms.

\section{Empirical Results}
\label{sec:exp}

We will now present empirical results illustrating the improvement in both variable-based and state-based identifiability due to CSI and state constraints, and further highlighting the separations between the two levels of identifiability. 

We consider the following setups of identifiability in our experiments: (1)~without constraints (\textsc{classical-id}), (2)~with CSI constraints only (\textsc{id-csi-heu}), and (3)~with both CSIs and state constraints, assuming all observed variables are binary (\textsc{id-csi-state-heu}). For all setups, we evaluate identifiability at both the variable-based level (denoted by the suffix ``\_var'') and state-based level (denoted by the suffix ``\_state''). Note that the two levels are equivalent in setup~(1) based on Proposition~\ref{prop:eq-classical}. For setup~(3), we also compare with the \textsc{dosearch} package~\citep{nips/TikkaHK19}, which targets variable-based identifiability for binary variables only.\footnote{The \textsc{dosearch} package is available at \url{https://github.com/santikka/dosearch}.}

For all sets of experiments, we randomly generate 200 causal graphs that contain $n \in \{6, 8, 10, 12, 15,$ $20, 30, 40, 50\}$ variables. Each variable has at most 5 parents. Among the variables, 70\% are designated to be observed, 20\% as treatment, and 20\% as outcome. The CSI constraints are randomly generated as follows. For each variable \(C\) with parent \(P_1\), we assign a CSI \((C \indep P_1\ |\ \p_2, \P_3)\) with probability $r \in \{0.2, 0.5, 0.8\}$.  We then assign a random state of other observed parents to \(\p_2\) with probability \(0.5.\) All remaining parents are assigned to \(\P_3.\) Note that larger values of $r$ correspond to more imposed CSIs.

Table~\ref{tab:exp1} in the Appendix reports the average number of identifiable causal effects and runtime for each method, with a 5-minute timeout for each run. Figure~\ref{fig:exp} plots the number of identifiable causal effects with an increasing number of variables $n$ and a fixed \(r\Equal 0.8\). The following patterns can be observed. First, for both variable-based and state-based identifiability, more causal effects become identifiable when additional CSI constraints are imposed (i.e., larger \(r\)) and when state constraints are present, which matches our intuition that constraints can facilitate identifiability. Second, the gap between state-based and variable-based identifiability widens as more CSI constraints are imposed. For example, when \(n\Equal 50\) and state constraints are present, the gap in the number of identifiable causal effects increases from $4$ to $16$ when $r$ increases from $0.2$ to $0.8$. A similar trend holds in the absence of state constraints. Finally, our algorithm \textsc{id-csi-state-heu} identifies a comparable -- sometimes larger -- number of variable-based causal effects compared to \textsc{dosearch}, while achieving significantly higher computational efficiency (\textsc{dosearch} times out on the majority of runs when $n \geq 15$). These results demonstrate the effectiveness of our proposed algorithms when applied to both state-based and variable-based identifiability.

It is worth emphasizing that our algorithms adopt a graph decomposition approach using c-components, in contrast to the exhaustive search method based on CSI-calculus rules introduced in~\citep{nips/TikkaHK19} and used in~\textsc{dosearch}. In fact, one may draw an interesting analogy here: our algorithms relate to CSI-calculus in a similar way to how \textsc{identify}~\citep{TianPearl03} relate to the do-calculus~\citep{pearl00b} as noted in~\citep{nips/TikkaHK19}. A key advantage of our procedural approach is that it makes the identification process more explicit than exhaustive search, enabling computational optimizations (e.g., the heuristics developed in Section~\ref{ssec:heuristics}) and facilitating future analysis of completeness. 

\section{Conclusion}
\label{sec:conclusion}
We considered how the identifiability of causal effects can be improved by focusing on a particular treatment and outcome and by including additional knowledge beyond causal graphs. This led us to
suggest the more granular notion of state-based identifiability. It also led to some findings on the extent to which various types of knowledge and their combinations can draw distinctions between state-based and variable-based identifiability. 
Another contribution of this work is the introduction of novel algorithms for both levels of causal effect identification under additional CSI and state constraints,
which we evaluated empirically.

\bibliography{main}

@book{pearl00b,
key = "pearl00b",
author = "Judea Pearl",
title = "Causality: Models, Reasoning, and Inference",
edition = "Second",
year = "2009",
publisher = "Cambridge University Press"
}

@Article{neurips/ChenDarwiche24,
AUTHOR = {Chen, Yizuo and Darwiche, Adnan},
TITLE = {Identifying Causal Effects Under Functional Dependencies},
JOURNAL = {Entropy},
VOLUME = {26},
YEAR = {2024},
NUMBER = {12},
ARTICLE-NUMBER = {1061},
PubMedID = {39766691},
ISSN = {1099-4300},
}

@article{pearl1993b,
 author = {Judea Pearl},
 journal = {Statistical Science},
 number = {3},
 pages = {266--269},
 publisher = {Institute of Mathematical Statistics},
 title = {[Bayesian Analysis in Expert Systems]: Comment: Graphical Models, Causality and Intervention},
 urldate = {2023-10-23},
 volume = {8},
 year = {1993}
}

@book{hernanbook,
    author = {Miguel A. Hernán and James M. Robins},
    title = {Causal Inference: What If},
    publisher = {Boca Raton: Chapman \& Hall/CRC},
    year = {2020}
}

@article{jmlr/ShpitserP08,
  author    = {Ilya Shpitser and
               Judea Pearl},
  title     = {Complete Identification Methods for the Causal Hierarchy},
  journal   = {J. Mach. Learn. Res.},
  volume    = {9},
  pages     = {1941--1979},
  year      = {2008}
}

@article{TianPearl03,
  author    = {Jin Tian and
               Judea Pearl},
  title     = {On the Identification of Causal Effects},
  journal   = {Technical Report},
  volume    = {R-290-L},
  year      = {2003}
}

@article{Verma93,
  author    = {Thomas S. Verma},
  title     = {Graphical Aspects of Causal Models},
  journal   = {Technical Report},
  volume    = {R-191},
  year      = {1993}
}

@inproceedings{aaai/ShpitserP06,
  author       = {Ilya Shpitser and
                  Judea Pearl},
  title        = {Identification of Joint Interventional Distributions in Recursive
                  Semi-Markovian Causal Models},
  booktitle    = {{AAAI}},
  pages        = {1219--1226},
  publisher    = {{AAAI} Press},
  year         = {2006}
}

@inproceedings{aaai/HuangV06,
  author       = {Yimin Huang and
                  Marco Valtorta},
  title        = {Identifiability in Causal Bayesian Networks: {A} Sound and Complete
                  Algorithm},
  booktitle    = {{AAAI}},
  pages        = {1149--1154},
  publisher    = {{AAAI} Press},
  year         = {2006}
}

@inproceedings{uai/BoutilierFGK96,
  author       = {Craig Boutilier and
                  Nir Friedman and
                  Mois{\'{e}}s Goldszmidt and
                  Daphne Koller},
  title        = {Context-Specific Independence in Bayesian Networks},
  booktitle    = {{UAI}},
  pages        = {115--123},
  publisher    = {Morgan Kaufmann},
  year         = {1996}
}

@inproceedings{pgm/ShenCD20,
  author       = {Yujia Shen and
                  Arthur Choi and
                  Adnan Darwiche},
  title        = {A New Perspective on Learning Context-Specific Independence},
  booktitle    = {{PGM}},
  series       = {Proceedings of Machine Learning Research},
  volume       = {138},
  pages        = {425--436},
  publisher    = {{PMLR}},
  year         = {2020}
}

@article{aict/ChenDarwiche25,
  author       = {Yizuo Chen and
                  Adnan Darwiche},
  title        = {Constrained Identifiability of Causal Effects},
  journal      = {AAAI Workshop on Artificial Intelligence with Causal Techniques},
  year         = {2025}
}

@inproceedings{uai/TianP02b,
  author       = {Jin Tian and
                  Judea Pearl},
  title        = {On the Testable Implications of Causal Models with Hidden Variables},
  booktitle    = {{UAI}},
  pages        = {519--527},
  publisher    = {Morgan Kaufmann},
  year         = {2002}
}

@inproceedings{uai/VermaP88,
  author       = {Thomas Verma and
                  Judea Pearl},
  title        = {Causal networks: semantics and expressiveness},
  booktitle    = {{UAI}},
  pages        = {69--78},
  publisher    = {North-Holland},
  year         = {1988}
}

@article{pearl95,
 ISSN = {00063444},
 author = {Judea Pearl},
 journal = {Biometrika},
 number = {4},
 pages = {669--688},
 publisher = {[Oxford University Press, Biometrika Trust]},
 title = {Causal Diagrams for Empirical Research},
 volume = {82},
 year = {1995}
}

@inproceedings{nips/TikkaHK19,
  author       = {Santtu Tikka and
                  Antti Hyttinen and
                  Juha Karvanen},
  title        = {Identifying Causal Effects via Context-specific Independence Relations},
  booktitle    = {NeurIPS},
  pages        = {2800--2810},
  year         = {2019}
}

@book{pearl18,
key = "pearl18",
author = "Judea Pearl and Dana Mackenzie",
title = "The Book of Why: The New Science of Cause and Effect",
year = "2018",
publisher = "Basic Books"
}

@article{datamine/PensarNKC15,
  author       = {Johan Pensar and
                  Henrik J. Nyman and
                  Timo Koski and
                  Jukka Corander},
  title        = {Labeled directed acyclic graphs: a generalization of context-specific
                  independence in directed graphical models},
  journal      = {Data Min. Knowl. Discov.},
  volume       = {29},
  number       = {2},
  pages        = {503--533},
  year         = {2015}
}

@inproceedings{aistats/MokhtarianJEK22,
  author       = {Ehsan Mokhtarian and
                  Fateme Jamshidi and
                  Jalal Etesami and
                  Negar Kiyavash},
  title        = {Causal Effect Identification with Context-specific Independence Relations
                  of Control Variables},
  booktitle    = {{AISTATS}},
  series       = {Proceedings of Machine Learning Research},
  volume       = {151},
  pages        = {11237--11246},
  publisher    = {{PMLR}},
  year         = {2022}
}

@inproceedings{uai/LeeGS24,
  author       = {Jaron Jia Rong Lee and
                  AmirEmad Ghassami and
                  Ilya Shpitser},
  title        = {A General Identification Algorithm For Data Fusion Problems Under
                  Systematic Selection},
  booktitle    = {{UAI}},
  series       = {Proceedings of Machine Learning Research},
  volume       = {244},
  pages        = {2188--2204},
  publisher    = {{PMLR}},
  year         = {2024}
}

\clearpage
\appendix
\section{Background on C-components and C-factors}
\label{app:ccomp}
Sound and complete methods for testing classical identifiability based on causal graphs have been developed in the past; see, e.g., \textsc{identify}~\citep{TianPearl03,aaai/HuangV06} and \textsc{id}~\citep{aaai/ShpitserP06}. These methods all rely on a key graphical notion called \emph{c-component}, which we review next.

\begin{definition}[c-component~\citep{uai/TianP02b}]
Let \(G\) be a causal graph with a subset of observed variables. A variable set \(\C\) is said to be a \underline{c-component} in \(G\) if it satisfies the following conditions: (i) every pair of hidden variables in \(\C\) are either adjacent or have a common child; and (ii) every observed variable in \(\C\) has some hidden parent in \(\C.\)
\end{definition}
A c-component is \emph{maximal} if it is not a subset of any other c-component. We can always \emph{partition} the variables (both observed and hidden) in a causal graph into maximal c-components. For example, Figure~\ref{sfig:ccomp3} depicts the c-component partition of the causal graph, where variables in different c-components are highlighted with different colors. 

Moreover, each c-component \(\C\) induces a \emph{c-factor} defined as \(Q[\C] = \sum_{\H} \prod_{C \in \C} \theta_{C | \P_C}\), where \(\H\) denotes the hidden variables in \(\C\) and \(\P_C\) denotes the parents of \(C\) in \(G.\) Based on the definition, the c-factor \(Q[\C]\) is over all observed variables appearing in the CPTs of \(\C.\) To illustrate, the c-factor for the c-component \(\{C, D, E, F\}\) in Figure~\ref{sfig:ccomp3} is computed as
\(Q[C DEF] = \sum_{CD}\theta_{C|A}\theta_{D | BC}\theta_{E|AC}\theta_{F|DE}\), which is over variables \(\{A, B, E, F\}.\)
One important property of c-factors is that they decompose the observational distribution \(\Pr(\V)\) as follows. Let \(\{\C_1, \dots, \C_m\}\) be the c-components that partition variables in \(G\), then \(\Pr(\V) = \prod_{i=1}^m Q[\C_i]\)~\citep{TianPearl03}. In Figure~\ref{sfig:ccomp3}, for example, \(\Pr(\V) = Q[A]Q[B]Q[CDEF]Q[GHI].\) In this work, we will apply this c-component (c-factor) decomposition technique for causal effect identification under constraints.

It is standard to convert the problem of identifying causal effects to independent problems of identifying c-factors as shown in the \textsc{identify} algorithm~\citep{TianPearl03}. In particular, a c-factor is identifiable iff its entries are uniquely determined (computable) by the observational distribution \(\Pr(\V).\)
Consider a causal graph \(G\), treatments \(\X\), and outcomes \(\Y.\) Let \(T\) be the mutilated graph obtained by first removing treatment variables \(\X\) from \(G\) and then retaining only the ancestors of \(\Y\) (Figure~\ref{sfig:ccomp2} depicts the mutilated graph of Figure~\ref{sfig:ccomp1}). The \textsc{identify} algorithm then reduces the identifiability of the causal effect \(\Pr_\X(\Y)\) to that of c-factors induced by the c-components in \(T\). This reduction is both sound and complete~\citep{aaai/HuangV06,aaai/ShpitserP06}: the causal effect is variable-based (state-based) identifiable iff all the c-factors are identifiable.

\begin{figure}[tb]
\centering
\begin{subfigure}[b]{0.32\linewidth}
\centering
\begin{tikzpicture}[->=stealth,auto,scale=\dagsize,transform shape]
\node[font=\huge] (A) at (0,0) {$A$};
\node[font=\huge] (B) at (4,0) {$B$};
\node[state,font=\huge] (C) at (0,-2.5) {$C$};
\node[state,font=\huge] (D) at (2.5,-2.5) {$D$};
\node[font=\huge] (E) at (0,-5) {$E$};
\node[font=\huge] (F) at (2.5,-5) {$F$};
\node[state,font=\huge] (G) at (5.5,-2.5) {$G$};
\node[font=\huge] (H) at (4.5,-5) {$H$};
\node[font=\huge] (I) at (6.5,-5) {$I$};

\path (A) edge (C);
\path (A) edge[bend right=30] (E);
\path (C) edge (E);
\path (B) edge (D);
\path (C) edge (D);
\path (E) edge (F);
\path (D) edge (F);
\path (B) edge (G);
\path (G) edge (H);
\path (G) edge (I);
\path (F) edge (H);
\path (H) edge (I);
\end{tikzpicture}
\caption{original $G$}
\label{sfig:ccomp1}
\end{subfigure}
\begin{subfigure}[b]{0.32\linewidth}
\centering
\begin{tikzpicture}[->=stealth,auto,scale=\dagsize,transform shape]
\node[font=\huge] (A) at (0,0) {$A$};
\node[font=\huge,blue] (B) at (4,0) {$B$};
\node[state,font=\huge,orange] (C) at (0,-2.5) {$C$};
\node[state,font=\huge,orange] (D) at (2.5,-2.5) {$D$};
\node[font=\huge,orange] (E) at (0,-5) {$E$};
\node[font=\huge,orange] (F) at (2.5,-5) {$F$};
\node[state,font=\huge,purple] (G) at (5.5,-2.5) {$G$};
\node[font=\huge,purple] (H) at (4.5,-5) {$H$};
\node[font=\huge,purple] (I) at (6.5,-5) {$I$};

\path (A) edge (C);
\path (A) edge[bend right=30] (E);
\path (C) edge (E);
\path (B) edge (D);
\path (C) edge (D);
\path (E) edge (F);
\path (D) edge (F);
\path (B) edge (G);
\path (G) edge (H);
\path (G) edge (I);
\path (F) edge (H);
\path (H) edge (I);
\end{tikzpicture}
\caption{c-components}
\label{sfig:ccomp3}
\end{subfigure}
\begin{subfigure}[b]{0.32\linewidth}
\centering
\begin{tikzpicture}[->=stealth,auto,scale=\dagsize,transform shape]
\node[font=\huge] (B) at (5.5,0) {$B$};
\node[state,font=\huge] (G) at (5.5,-2.5) {$G$};
\node[font=\huge] (H) at (4.5,-5) {$H$};
\node[font=\huge] (I) at (6.5,-5) {$I$};

\path (B) edge (G);
\path (G) edge (H);
\path (G) edge (I);
\path (H) edge (I);
\end{tikzpicture}
\caption{mutilated}
\label{sfig:ccomp2}
\end{subfigure}
\caption{The causal graph \(G\), its mutilated graph under \(\Pr_{F}(I)\), and the c-component decomposition. Hidden variables are shown in rectangles, and variables in different c-components are highlighted with different colors.}
\label{fig:ccomp}
\end{figure}

\section{Properties of Context-Induced Graphs}
\label{sec:ci-graph}
Context-induced graphs (CI-graphs) introduced in Definition~\ref{def:ci-graph} satisfy two key properties which we formalize in the following proposition.
\begin{proposition}
\label{prop:ci-graph}
Consider a causal graph \(G\) and a set of CSIs \(\CC\). Let \(T\) be a CI-graph under context \(\s\). Then the following properties hold for all models \(\Theta\) of \(G\) satisfying \(\CC\):
\begin{enumerate}
\item[(i)] Let \(\X, \Y, \Z, \S\) be disjoint sets of variables. If \(\X\) and \(\Y\) are d-separated by \((\Z \cup \S)\) in \(T\), then \((\X \indep \Y\ |\ \Z, \s)\) in the distribution induced by \(\Theta.\)\footnote{See~\citep{uai/VermaP88} for details of d-separation and its interplay with conditional independencies.}
\item[(ii)] For each variable \(V\) with parents \(\P\) in \(G\) and parents \(\P'\) in \(T\), \(\theta_{v | \p_1} = \theta_{v | \p_2}\) for all instantiations \(\p_1, \p_2\) of \(\P\) that extend a common \(\p'\) that is compatible with \(\s\).\footnote{\label{ft:ci-graph}Let \(\x, \y\) be two instantiations. We say that \(\x\) \emph{extends} \(\y\) if \(\x\) contains all the states in \(\y\), and that \(\x\) and \(\y\) are \emph{compatible} if they do not contain conflicting states. The CI-graphs considered here differ from the notion of \emph{context-specific graph} in~\citep{datamine/PensarNKC15,nips/TikkaHK19,uai/LeeGS24}: the latter satisfies property~(i) but not property~(ii). Moreover, unlike context-specific graphs, CI-graphs are not necessarily unique.}
\end{enumerate}
\end{proposition}

The first property shows that CI-graphs preserve the conditional independencies given the context \(\s\), and the second property implies that the parameters in CI-graphs characterize the equal parameters under the CSIs. In particular, we can always construct a model for the CI-graph by setting each parameter \(\theta_{v | \p'}\) equal to \(\theta_{v | \p}\) for any \(\p \supseteq \p'\) (the value of \(\theta_{v | \p'}\) is unique if \(\p'\) is compatible with \(\s\)). The resulting model satisfies several important properties, including the preservation of \(\Pr(\v)\) for all \(\v\) compatible with the context \(\s\). We will use this construction technique to prove the soundness of our identification method (Algorithm~\ref{alg:id-csi}).

\begin{wrapfigure}[4]{r}{0.2\linewidth} 
\centering
\begin{tikzpicture}[->=stealth,auto,scale=\dagsize,transform shape]
\node[state,font=\huge] (U) at (0,0) {$A$};
\node[font=\huge] (Y) at (1,-2) {$D$};
\node[font=\huge] (X1) at (-1,-2) {$B$};
\node[font=\huge] (X2) at (2,0) {$C$};
\node[font=\huge] (E) at (3,-2) {$E$};

\path (U) edge (Y);
\path (U) edge (X1);
\path (X1) edge (Y);
\path (X2) edge (E);
\path (X1) edge[bend right=20] (E);
\end{tikzpicture}
\label{fig:wrap-right}
\end{wrapfigure}We next illustrate the properties in Proposition~\ref{prop:ci-graph} using
the following example. The figure on the right depicts a CI-graph
of Figure~\ref{sfig:prune1} under context \(b_0\), assuming CSIs \((D \indep B\ |\ c_0, A)\), \((D \indep C\ |\ b_0, A),\) and \((E \indep B\ |\ c_0).\) Since \(A\) and \(E\) are d-separated by \(B\) in the CI-graph, \((A \indep E\ |\ b_0)\) by property~(i). Moreover, \(\theta_{D | A, b_0, c_1} = \theta_{D | A, b_0, c_2}\) holds for any states \(c_1, c_2\) of \(C\) by property~(ii).

\section{Heuristics} 
\label{app:heuristics}
We present heuristics that select a single choice for the CI-graph under \(\x\y\) (Algorithm~\ref{alg:id-csi}, line~2), the context \(\s\) and its CI-graph (Algorithm~\ref{alg:id-csi}, lines~6-7), and the context-partition (Algorithm~\ref{alg:id-csi-state} line~2) when state constraints are imposed. Although the use of heuristics may compromise the completeness of the algorithms, it makes them computationally feasible by avoiding exponential enumerations and still substantially improves the identifiability of causal effects, as demonstrated by the empirical results in Section~\ref{sec:exp}.

\paragraph{CI-graph under \(\x\y\) (Algorithm~\ref{alg:id-csi}, line~2).} We begin with a reverse-topological ordering \(\Pi\) over states \(\x \y.\) Let \(\s\) be a context initialized to be empty, and \(T\) be a graph initialized as the original graph \(G\). We iterate over each state in \(\Pi\) in order, adding it to \(\s\) and removing from \(T\) all edges enabled by the CSIs under the updated \(\s\). The resulting graph \(T\) is guaranteed to be a CI-graph for \(\x\y\), and no additional edges can be pruned from the CI-graph after the procedure.

\paragraph{Context \(\s\) and its CI-graph \(H\) (Algorithm~\ref{alg:id-csi}, lines~6-7).} The intuition is to activate CSIs and remove edges one at a time, returning immediately if the CI-graph can be used to establish identifiability of the target c-component $\C_i$.

We initialize the context \(\s\) as follows. For each variable \(W\) that is in \(\C_i\) or is a parent of some variable in \(\C_i\) and has a state \(w \in \x\y\), we add \(w\) to \(\s.\) We initialize \(H\) as the original graph \(G\). For each variable \(C \in \C_i\) with some parent \(P\), we remove the edge \(P \rightarrow C\) from \(H\) if the edge is not in \(T\). The initialized \(\s\) and \(H\) satisfy the condition in line~7.

We now augment the context \(\s\) and prune additional edges from \(H\) in an iterative way.
At each iteration, let \(\C \supseteq \C_i\) be the c-component of \(H\) resulting from \textsc{id-qfunc}. If \(\C = \C_i,\) we terminate the iterative process since the c-factor of \(\C_i\) is identifiable under \(\s\) and \(H.\) Otherwise, we pick a CSI \((W \indep P_1\ |\ \p_2, \P_3)\) that satisfies the following conditions: \(W \in (\C \setminus \C_i)\), \(P_1 \in \C,\) and \(\p_2\) is compatible with \(\s.\) When multiple such CSIs exist, we prioritize the one whose \(P_1\) is hidden and has fewest children in \(H_{\C},\) where \(H_{\C}\) denotes the subgraph of \(H\) only consisting of \(\C.\) We then add states \(\p_2\) to the context \(\s\) and remove the edge \(P_1 \rightarrow C\) from \(H,\) before proceeding to the next iteration. The process terminates if no CSI satisfies the above conditions.

\paragraph{Context-partition (Algorithm~\ref{alg:id-csi-state} line~2).} The intuition is to include as few variables as possible in the context-partitions, as long as doing so is unlikely to compromise identifiability. In particular, we construct the contexts in the partition and their corresponding CI-graphs simultaneously. We begin with a reverse-topological ordering \(\Pi\) over observed variables \(\V.\) Let \(\s\) be a context initialized to be empty, and \(T\) be a graph initialized as the original graph \(G\). 

We iterate over each variable \(V\) in \(\Pi\) in order. At each iteration, let \(M\) be the mutilated graph obtained by removing \(\X\) and retaining the ancestors of \((\Y\cup \S)\) in \(T.\) If \(V \in \X \cup \Y\) with state \(v \in \x\y\), we remove from \(T\) all edges enabled by the CSIs under the context \(\s \cup \{v\}\), and continue the iteration using the current context \(\s\) and the pruned graph. 

If \(V \notin \X \cup \Y\) but every state of \(V\) appears in the context of some CSI in $M$, we create a new context \(\s \cup \{v_i\}\) for each state \(v_i\) of \(\V\). For each new context, we assign a graph obtained by removing from \(T\) all edges enable by the CSIs under that new context. We then continue the iteration for each new context and its assigned graph.

Otherwise, we skip \(V\) and proceed with the current context \(\s\) and graph \(T.\)
After all variables have been processed, we collect the resulting contexts to form the context-partition and take their associated graphs as the CI-graphs.

\section{Statements and Proofs}
\label{app:proof}
The statements and proofs are presented in a different order from that of the main paper. We begin with the soundness proofs of the algorithms introduced in Section~\ref{sec:main-algs} and then establish the separation results in Sections~\ref{sec:csi} and~\ref{sec:domain-constraints}.
\subsection{Soundness Proofs}
\begin{proof}[Proof of Proposition~\ref{prop:ci-graph}]
We first show that property~(ii) holds after each edge removal in the definition of CI-graph (Definition~\ref{def:ci-graph}). Initially, \(\P' = \P\) and hence \(\theta_{v | \p_1} = \theta_{v | \p_2}\) for \(\p_1 = \p_2.\) Suppose now the property holds for graph \(T\) before the removal of edge \(P_1 \rightarrow C\) based on the CSI \((C \indep P_1\ |\ \p_2, \P_3)\). We want to show that property~(ii) holds after the edge removal. Let \(\P\) be the parents of \(C\) in \(T\) and \(\P'\) be the parents after removal. Let \(\p^1, \p^2\) be any two instantiations extending a common \(\p'\) compatible with \(\s.\) 

Suppose \(C\) appears in the context, i.e., \(c^\star \in \s\) for some state \(c^\star\). Then \(\theta_{v | \p^1} = \theta_{v | \p^1_{c^\star}}\) and \(\theta_{v | \p^2} = \theta_{v | \p^2_{c^\star}}\) according to the CSI constraint, where \(\p^1_{c^\star}\), \(\p^2_{c^\star}\) are obtained from \(\p_1, \p_2\) by replacing the state of \(C\) by \(c^\star.\) It follows that \(\p^1_{c^\star}\), \(\p^2_{c^\star}\) extend a common \(\p\) that is compatible with \(\s.\) Hence, \(\theta_{v | \p^1} = \theta_{v | \p^1_{c^\star}} = \theta_{v | \p^2_{c^\star}} = \theta_{v | \p^2}\) by induction.

Suppose \(C\) does not appear in the context, i.e., \(C \notin \S.\) Let \(c^\star\) be an arbitrary state of \(C.\) Then \(\theta_{v | \p^1} = \theta_{v | \p^1_{c^\star}}\) and \(\theta_{v | \p^2} = \theta_{v | \p^2_{c^\star}}\) according to the CSI constraint. Since both \(\p^1_{c^\star}\) and \(\p^2_{c^\star}\) are compatible with \(\s,\) \(\theta_{v | \p^1} = \theta_{v | \p^1_{c^\star}} = \theta_{v | \p^2_{c^\star}} = \theta_{v | \p^2}\) by induction.

We next prove property~(i). For any model \(\Theta\) for \(G\) that induces \(\Pr,\) we want to construct a model for \(T\) that induces a distribution \(\Pr'\) satisfying \(\Pr(\v) = \Pr'(\v)\) for all instantiations \(\v \supseteq \s.\) This ensures that if d-separation holds, i.e., \(\Pr'(\X | \Y, \Z, \s) = \Pr'(\X | \Z, \s)\), the conditional independency also holds, i.e., \(\Pr(\X | \Y, \Z, \s) = \Pr(\X | \Z, \s)\). The construction can be done as follows. For each variable \(C\) with parents \(\P'\) in \(T\), we define its CPT entries as \(\theta'_{c|\p'} = \theta_{c | \p}\) for every instantiation \((c, \p')\), where \(\p\) can be any instantiation where \(\p \supseteq \p'.\) By property~(ii), when \(\p'\) is compatible with \(\s\), the value of \(\theta'_{c | \p'}\) is independent of the choice of \(\p\) since all the \(\theta_{c | \p}\) are equal. Hence, \(\Pr(\v) = \Pr'(\v)\) for all \(\v\) compatible with \(\s.\)
\end{proof}

Before showing the soundness proofs for the algorithms, we first clarify the models (parameterizations) for CI-graphs so that the distributions induced by CI-graphs are well-defined. Let \(T\) be a CI-graph of the original graph \(G\) under some context \(\s\) by Definition~\ref{def:ci-graph}. We assume that the model \(\Theta'\) of \(T\) is constructed from the model \(\Theta\) of \(G\) as follows. For each variable \(C\) with parents \(\P\) in \(G\) and parents \(\P'\) in \(T\), we assign \(\theta'_{c | \p'} = \theta_{c | \p}\) for every instantiation \((c,\p')\), where \(\p\) can be any instantiation satisfying \(\p \supseteq \p'\); this is the same as the construction method shown in Appendix~\ref{sec:ci-graph}. Based on the construction, we can now establish some key results on the c-factors induced by the CI-graph.
\begin{lemma}
\label{thm:id-ccomp}
Consider a CI-graph \(T\) under the context \(\x\y\). Let \(M\) be the mutilated graph constructed in Algorithm~\ref{alg:id-csi} line~3 and let \(\{\C_1, \dots, \C_m\}\) denote its maximal c-components. The causal effect \(\Pr_\x(\y)\) is identifiable if the c-factors \(Q[\C_i]_{\x\y}\) in \(T\) are identifiable for all \(i=1, \dots, m.\)
\end{lemma}
\begin{proof}
Let \(\Theta\) be the model for the original graph \(G\) and \(\Theta'\) be the model for the CI-graph \(T\) constructed using the above procedure.
Let \(\Pr\) and \(\Pr'\) be the distributions induced by \(\Theta\) and \(\Theta'\), respectively. By Proposition~\ref{prop:ci-graph}, \(\Theta'\) agrees with \(\Theta\) on all CPT entries that are compatible with \(\x\y\). As a result, \(\Pr'_\x(\v) = \Pr_\x(\v)\) for all instantiations \(\v \supseteq \y\). Therefore, the CI-graph \(T\) can be used for causal identification since it preserves the value of causal effect, i.e., \(\Pr'_\x(\y) = \Pr_\x(\y)\). The techniques of graph mutilation and c-component (c-factor) decomposition follow from classical identification methods such as \textsc{identify}~\citep{TianPearl03}.
\end{proof}

\begin{figure}[tb]
\centering
\begin{subfigure}[b]{0.49\linewidth}
\centering
\begin{tikzpicture}[->=stealth,auto,scale=\dagsize,transform shape]
\node[state,font=\huge] (U) at (0,0) {$U$};
\node[font=\huge] (X) at (-2,-2) {$X$};
\node[font=\huge] (Z) at (0,-2) {$Z$};
\node[font=\huge] (Y) at (2,-2) {$Y$};

\path (U) edge (X);
\path (X) edge (Z);
\path (U) edge (Z);
\path (U) edge (Y);
\path (Z) edge (Y);
\end{tikzpicture}
\caption{Original \(G\)}
\label{sfig:count-eg1}
\end{subfigure}
\begin{subfigure}[b]{0.49\linewidth}
\centering
\begin{tikzpicture}[->=stealth,auto,scale=\dagsize,transform shape]
\node[state,font=\huge] (U) at (0,0) {$U$};
\node[font=\huge] (X) at (-2,-2) {$X$};
\node[font=\huge] (Z) at (0,-2) {$Z$};
\node[font=\huge] (Y) at (2,-2) {$Y$};

\path (U) edge (X);
\path (X) edge (Z);
\path (U) edge (Y);
\path (Z) edge (Y);
\end{tikzpicture}
\caption{CI-graph \(T\)}
\label{sfig:count-eg2}
\end{subfigure}
\caption{CI-graph constructed for the causal effect \(\Pr_x(y)\) and CSI constraint \((Z \indep U\ |\ x).\)}
\label{fig:count-eg}
\end{figure}

\begin{proposition}
\label{prop:incorrect-ci-graph}
There exists a graph \(G\) and its CI-graph \(T\) under context \(\x\y\) for which \(\Pr_\x(\y)\) is unidentifiable wrt \(\langle G, \V, \CC\rangle\) but identifiable wrt \(\langle T, \V \rangle.\)
\end{proposition}
\begin{proof}
Consider the causal graph \(G\) in Figure~\ref{sfig:count-eg1} with the CSI constraint \((Z \indep U\ |\ x).\) To compute the causal effect \(\Pr_x(y)\), we construct the CI-graph \(T\) under \(xy\) as shown in Figure~\ref{sfig:count-eg2}. Applying classical identification methods (e.g., the front-door criterion~\cite{pearl95}) to \(T\) would suggest that the causal effect is identifiable. However, this conclusion is incorrect: we can construct the following parameterizations \(\Theta^1, \Theta^2\) for \(G\) that satisfy the CSI constraint and induce the same \(\Pr(\V)\) yet yield distinct \(\Pr_x(y).\)

\begin{table}[H]
\centering
\begin{tabular}{c|c|c}
$U$ & \(\theta^1_U\) & \(\theta^2_U\)\\
\hline
$u$ & $0.5$ & $0.4$ \\
$\bar{u}$ & $0.5$ & $0.6$\\ 
\end{tabular} 
\begin{tabular}{c|c|c|c}
$U$ & $X$ & \(\theta^1_{X|U}\) & \(\theta^2_{X|U}\)\\
\hline
$u$ & $x$ & $0.6$ & $0.25$ \\
${u}$ & $\bar{x}$ & $0.4$ & $0.75$\\ 
\hline
$\bar{u}$ & $x$ & $0.2$ & $0.5$ \\
$\bar{u}$ & $\bar{x}$ & $0.8$ & $0.5$\\ 
\end{tabular} 

\begin{tabular}{c|c|c|c|c}
$U$ & $X$ & $Z$ & \(\theta^1_{Z | U,X}\) & \(\theta^2_{Z | U,X}\)\\
\hline
$u$ & $x$ & $z$ & $0.5$ & $0.5$ \\
$u$ & $x$ & $\bar{z}$ & $0.5$ & $0.5$ \\
$u$ & $\bar{x}$ & $z$ & $0.3$ & $1/3$ \\
$u$ & $\bar{x}$ & $\bar{z}$ & $0.7$ & $2/3$ \\
\hline
$\bar{u}$ & $x$ & $z$ & $0.5$ & $0.5$ \\
$\bar{u}$ & $x$ & $\bar{z}$ & $0.5$ & $0.5$ \\
$\bar{u}$ & $\bar{x}$ & $z$ & $0.6$ & $2/3$ \\
$\bar{u}$ & $\bar{x}$ & $\bar{z}$ & $0.4$ & $1/3$ \\
\end{tabular}

\begin{tabular}{c|c|c|c|c}
$U$ & $Z$ & $Y$ & \(\theta^1_{Y|U,Z}\) & \(\theta^2_{Y|U,Z}\)\\
\hline
$u$ & $z$ & $y$ & $0.5$ & $0.76$ \\
$u$ & $z$ & $\bar{y}$ & $0.5$ & $0.24$ \\
$u$ & $\bar{z}$ & $y$ & $0.8$ & $0.42$ \\
$u$ & $\bar{z}$ & $\bar{y}$ & $0.2$ & $0.58$ \\
\hline
$\bar{u}$ & $z$ & $y$ & $0.55$ & $0.43$ \\
$\bar{u}$ & $z$ & $\bar{y}$ & $0.45$ & $0.57$ \\
$\bar{u}$ & $\bar{z}$ & $y$ & $0.3$ & $0.76$ \\
$\bar{u}$ & $\bar{z}$ & $\bar{y}$ & $0.7$ & $0.24$\\
\end{tabular} 
\end{table}
\end{proof}

\begin{lemma}
\label{thm:cc-id}
Let \(T\) be a CI-graph of the original graph \(G\) under some context \(\s.\) For each c-component \(\C\) in \(T\) (not necessarily maximal), its corresponding c-factor \(Q[\C]_\s\) is identifiable if \(\C\) can be obtained from \(T\) by repeatedly applying the following operations:\footnote{For simplicity, we use \(T\) to refer to the graph and its set of variables interchangeably.}
\begin{enumerate}
\item[(i).] Remove a leaf node that is not in \(\S\); or
\item[(ii).] Replace \(T\) with the subgraph formed by one of its maximal c-components.
\end{enumerate}
\end{lemma}
\begin{proof}
Consider again the construction method above Lemma~\ref{thm:id-ccomp}.
Let \(\Theta\) and \(\Theta'\) be the parameterizations for the original graph \(G\) and CI-graph \(T\). Let \(\Pr\) and \(\Pr'\) be the distributions induced by \(\Theta\) and \(\Theta'\). By Proposition~\ref{prop:ci-graph} property~(ii), the equality \(\Pr(\u,\v) = \Pr'(\u, \v)\) holds for all instantiations \((\v, \u)\) compatible with \(\s\).

We first extend the definition of c-factors to any variable sets \(\W\) (not necessarily c-components):
\(Q[\W] = \sum_{\U_\W} \prod_{W \in \W}\theta_{W | \P_W},\) where \(\U_\W\) are the hidden variables in \(\W\) and \(\P_W\) contains the parents of \(W\) in \(T.\) When \(\W\) is a c-component in \(T\), the definition is identical to the one in Appendix~\ref{app:ccomp}. Moreover, we use \(Q[\W]_\s\) to denote the instantiations of \(Q[\W]\) compatible with the context \(\s.\)

Our goal is to show that the c-factors obtained from the two operations in this lemma are identifiable (computable from \(\Pr(\V)\)). 
Let \(\W\) be the result of applying the two operations. We prove the following statements by induction: (i) \(Q[\W]_{\s}\) is identifiable; and (ii) \(Q[\W^{(i)}]_{\s}\) is identifiable for each observed \(W^i \in \W\), where \(W^{i}\) is the $i^{th}$ variable in \(\W\) based on some fixed topological order \(\Pi\) of \(G,\) \(\W^{(i)}\) denotes all variables ordered before and including \(W^i\), and \(Q[\W^{(i)}]\) is defined as \(\sum_{\W \setminus \W^{(i)}}Q[\W]\)~\cite{uai/TianP02b}.

We start with the base case where \(\W\) contains all variables in \(T\), i.e., \(\W = \V \cup \U\). By construction, \(Q[\W]_\s = \Pr'(\V)_\s = \Pr(\V)_\s\) and hence statement~(i) holds. Now consider statement~(ii). For each observed \(W^i \in \W\), all variables in \((\W \setminus \W^{(i)})\) can be pruned from both \(G\) and \(T\). The CPT entries for the remaining variables \(\W^{(i)}\) agree on all instantiations compatible with \(\s.\) Hence, \(Q[\W^{(i)}]_\s = \Pr'(\W^{(i)} \cap \V)_\s = \Pr(\W^{(i)} \cap \V)_\s\) which is identifiable.

Suppose both statements hold for some variable set \(\W.\) Suppose a leaf \(L \notin \S\) is pruned from \(\W\). Let \(\D\) be the remaining variables, then the resulting c-factor \(Q[\D]_\s\) is
\[Q[\D]_\s = (\sum_L Q[\W])_\s = \sum_L Q[\W]_\s\]
By induction, the identifiability of \(Q[\D]_\s\) follows from the identifiability \(Q[\W]_\s\).

We next show that \(Q[\D^{(i)}]\) is identifiable for each observed \(D^{i} \in \D\) ordered based on \(\Pi\). WLG, suppose \(D^i\) is ranked as the $j^{th}$ variable in \(\W\), i.e., \(D^i=W^j.\) We consider two cases based on the location of the pruned leaf \(L\). If \(L \in \W^{(j)}\), then
\[Q[\D^{(i)}]_\s = (\sum_L \sum_{\W \setminus \W^{(j)}} Q[\W])_{\s} = \sum_L Q[\W^{(j)}]_{\s}\]
which can be computed by summing out \(L\) from \(Q[\W^{(j)}]_{\s}.\)

If \(L \notin \W^{(j)},\) then 
\[Q[\D^{(i)}]_\s = (\sum_{\W \setminus \W^{(j)}} Q[\W])_{\s} = Q[\W^{(j)}]_\s\]
which is also equivalent to summing out \(L\) from \(Q[\W^{(j)}]_\s.\) Since \(Q[\W^{(j)}]_\s\) is identifiable by induction, \(Q[\D^{(i)}]_\s\) is also identifiable in both cases.

Suppose now \(\W\) is replaced by a maximal c-component \(\D \subset \W\). According to~\cite[Lemma~4]{TianPearl03}, the c-factor for \(\D\) can be computed as \(Q[\D] =\) \(\prod_{W^i \in \D \cap \V}\) \(\frac{Q[\W^{(i)}]}{Q[\W^{(i-1)}]}.\) Since each \(Q[\W^{(i)}]_{\s}\) is identifiable by induction, \(Q[\D]_{\s}\) is also identifiable. We next show that \(Q[\D^{(i)}]_{\s}\) is identifiable for each observed \(D^i \in \D.\) By definition, \(Q[\D^{(i)}] = \sum_{\D \setminus \D^{(i)}} Q[\D]\). To simplify the expression further, we iteratively eliminate observed variables in (\(\D \setminus \D^{(i)}\)) in reverse of \(\Pi\) (we do not need to eliminate hidden variables since they have already been summed out in \(Q[\D]\)). WLG, let \(W^j\) be the first observed variable in the reverse order, then
\begin{equation*}
\begin{split}
&\sum_{W^j} Q[\D] = \sum_{W^j}\prod_{W^k \in \D \cap \V}\frac{Q[\W^{(k)}]}{Q[\W^{(k-1)}]} \\
&= (\prod_{W^k \in \D \cap \V, k < j} \frac{Q[\W^{(k)}]}{Q[\W^{(k-1)}]}) (\sum_{W^j} \frac{Q[\W^{(j)}]}{Q[\W^{(j-1)}]}) \\
&= \prod_{W^k \in \D \cap \V, k < j} \frac{Q[\W^{(k)}]}{Q[\W^{(k-1)}]}
\end{split}
\end{equation*}
The second equality holds since \(W^j\) appears only in \(\W^{(j)}\). We can apply the same procedure to eliminate all variables in \((\D \setminus \D^{(i)})\), yielding \(Q[\D^{(i)}] =\) \(\prod_{W^k \in \D^{(i)} \cap \V}\) \(\frac{Q[\W^{(k)}]}{Q[\W^{(k-1)}]}.\) Since each \(Q[\W^{(k)}]_\s\) is identifiable by induction, \(\prod_{W^k \in \D^{(i)} \cap \V}\) \(\frac{Q[\W^{(k)}]_\s}{Q[\W^{(k-1)}]_\s}\) is also identifiable. 
\end{proof}

\begin{proof}[Proof of Theorem~\ref{thm:sound}]
Suppose Algorithm~\ref{alg:id-csi} returns an identifying formula, there exists a CI-graph \(T\) under context \(\x\y\) such that the c-factors in the mutilated graph are deemed identifiable (lines~6-9). By Lemma~\ref{thm:id-ccomp}, it suffices to show that a c-factor is indeed identifiable if \textsc{id-qfunc} returns an identifying formula for some context \(\s\) and \(H\) in line~6.

Since \(\s\) and \(H\) satisfy the condition in line~7, the c-factor \(Q[\C_i]_{\x\y}\) in \(T\) can be attained from the c-factor \(Q[\C_i]_{\s}\) in \(H\) by selecting instantiations compatible with \(\x\y.\) This is because the two c-factors involve the same set of CPTs, and the context \(\s\) does not exclude any instantiations beyond those ruled out by the context \(\x\y\). Hence, \(Q[\C_i]_{\x\y}\) is identifiable if \(Q[\C_i]_{\s}\) is identifiable. Since the identification method \textsc{id-qfunc} follows the two operations described in Lemma~\ref{thm:cc-id}, the c-factor \(Q[\C_i]_{\s}\) is identifiable if \textsc{id-qfunc} returns a formula for its estimation by the lemma.
\end{proof}

\begin{proof}[Proof of Proposition~\ref{prop:id-complete}]
When no CSI is imposed, i.e., \(\CC = \emptyset,\) the CI-graph is identical to the original causal graph \(G.\) By setting \(\s=\emptyset\) and \(H = G\) in line~6, Algorithm~\ref{alg:id-csi} becomes identical to the \textsc{identify} algorithm in~\citep{TianPearl03}. The operation of pruning leaves from \(H_{\D}\) in lines~16-19 subsumes the operation of keeping the ancestors of \(\C\) in \textsc{identify}, and the operation of finding a maximal c-component in lines~20-22 is identical to the one in \textsc{identify}. Since \textsc{identify} has been shown to be sound and complete~\citep{aaai/HuangV06}, Algorithm~\ref{alg:id-csi} inherits the same properties for classical identifiability.
\end{proof}

\begin{proof}[Proof of Theorem~\ref{thm:partition}]
Suppose there is a context-partition \(\{\s_1, \dots, \s_k\}\) over \(\V \setminus (\X \cup \Y)\) such that each \(\Pr_\x(\y, \s_i)\) is identifiable by Algorithm~\ref{alg:id-csi}. Then \(\Pr_\x(\y) = \sum_{i=1}^k\) \(\Pr_\x(\y, \s_i)\) is identifiable by the law of total probability.
\end{proof}

\subsection{Separation Results}
\begin{proof}[Proof of Proposition~\ref{prop:eq-classical}]
The only-if direction directly follows from the definitions. We now show the contrapositive of the if direction. Suppose \(\Pr_\X(\Y)\) is unidentifiable, there exist some \(\x'\) and \(\y'\) such that \(\Pr_{\x'}(\y')\) is unidentifiable. We can simply set \(\x=\x'\) and \(\y=\y'\) for the states in \(\Pr_\x(\y).\)
\end{proof}

\begin{proof}[Proof of Proposition~\ref{prop:state-var-eq}]
If \(\Pr_\x(\y)\) is unidentifiable, then \(\Pr_\X(\Y)\) is also unidentifiable by the definition of variable-based identifiability. Now consider the opposite direction. If \(\Pr_\X(\Y)\) is unidentifiable, then there exist states \(\x\) and \(\y\) such that \(\Pr_\x(\y)\) is unidentifiable. We go over each state \(s\) in \(\x\y\) and create a new state \(s'\) (outside CSI contexts) for \(S\). The CPTs for \(S\) and each of its children \(C\) are updated as follows:
\[\theta'_{s' | \p} = \theta_{s | \p} \cdot \epsilon,\ \ \ \theta'_{s | \p} = \theta_{s | \p} \cdot (1 - \epsilon),\ \ \theta'_{c | \p_{C},s'} = \theta_{c | \p_{C}, s}\]
where \(\P_C\) denotes the parents of \(C\) (excluding \(S\)) and \(\epsilon\) is a positive constant in range \((0,1)\). The construction ensures that all CSIs satisfied by the original parameterization \(\Theta\) are also satisfied by the new parameterization \(\Theta'.\)

Let \(\x'\) and \(\y'\) be the results of replacing \(s\) by \(s'\) in \(\x\) and \(\y.\) Then the update parameterization induces the following observational distribution \(\Pr'(\V)\) and causal effect \(\Pr'_{\x'}(\y')\):
\[\Pr'(\v) = \begin{cases}
  \Pr(\v|_{s}) \cdot \epsilon  & s' \in  \v \\
  \Pr(\v) \cdot (1-\epsilon) & s \in \v \\
  \Pr(\v) & \text{otherwise}\\
\end{cases}
\]
\[\Pr'_{\x'}(\y') = \begin{cases}
  \Pr_\x(\y) \cdot \epsilon & S \in \Y\\
  \Pr_\x(\y)   & \text{otherwise}\\
\end{cases}
\]
where \(\v|_{s}\) replaces the state of \(S\) in \(\v\) by the state \(s.\)
Hence, if there exist parameterizations \(\Theta^1\), \(\Theta^2\) that induce a same \(\Pr(\V)\) yet distinct \(\Pr_\x(\y)\), we can apply the above procedure to construct parameterizations that induce a same \(\Pr'(\V)\) yet distinct \(\Pr_{\x'}(\y')\). By iterating over all states in \(\x\y,\) we conclude that there exist \(\x',\y'\) that do not involve any contexts of CSIs and for which \(\Pr_{\x'}(\y')\) is unidentifiable. We finally set \(\x=\x'\) and \(\y=\y'\) for \(\Pr_\x(\y).\)
\end{proof}

\begin{proposition}
\label{prop:cid-diff}
There exists a tuple \(\langle  G, \V, \CC \rangle\) where $\CC$ contains only CSI constraints for which \(\Pr_\X(\Y)\) is unidentifiable but \(\Pr_\x(\y)\) is identifiable for some states \(\x,\y.\)
\end{proposition}
\begin{proof}
Consider the causal graph in Figure~\ref{sfig:csi-eg1}. Let ``entry-level'' be state 0 of $J$, ``\(Y \geq 10\)'' be state 0 of \(Y\), and ``low'' be state 0 of $S.$ The CSI constraint can be written as (\(S \indep D | J=0, Y\)). 

We apply Algorithm~\ref{alg:id-csi} to show the identifiability of \(\Pr_{Y=0}(J=0,S=0).\) Figure~\ref{sfig:ci-mut-eg12} depicts the CI-graph \(T\) under the context \(\{Y=0, J=0, S=0\}\) and Figure~\ref{sfig:ci-mut-eg13} depicts its mutilated graph. We are left to show that the c-factors  \(Q[ADJ]_{Y=0,J=0,S=0}\) and \(Q[S]_{Y=0,J=0,S=0}\) in the mutilated graph are identifiable. 

To identify the first c-factor, we consider the context \(\s:\{J=0\}\) and its corresponding CI-graph which coincides with \(T.\) Since the context and CI-graph satisfy the condition in line~7 of Algorithm~\ref{alg:id-csi}, we can apply the \textsc{id-qfunc} procedure to establish the identifiability of \(Q[ADJ]_{J=0}.\) This is because the c-component \(\{A,D,J\}\) can be obtained from CI-graph \(T\) by first finding the maximal c-component \(\{A,D,J,Y\}\) and then pruning the leaf node \(Y.\) To identify the second c-factor, we again consider the context \(\s:\{J=0\}\) and its CI-graph depicted in \(T.\) Since \(\{S\}\) is directly a maximal c-component in \(T,\) the c-factor \(Q[S]_{J=0}\) is identifiable according to \textsc{id-qfunc}. Together, the causal effect  \(\Pr_{Y=0}(J=0,S=0)\) is identifiable since the entries in c-factors \(Q[ADJ]_{Y=0,J=0,S=0}\) and \(Q[S]_{Y=0,J=0,S=0}\) can be extracted from \(Q[ADJ]_{J=0}\) and \(Q[S]_{J=0}.\)

We next show that the variable-based causal effect \(\Pr_Y(J, S)\) is not identifiable. We assume that \(J\) has three states \(\{0, 1, 2\}\) and all other variables are binary with states \(\{0,1\}.\) For both parameterizations \(\Theta_1, \Theta_2\), we assign uniform distributions for \(A, D\), structural equation \(Y = A,\) and the following CPT for \(J\): 
\[\theta^1_{J | A, D} = \theta^2_{J | A, D} = \begin{cases}
0.01 & \text{if \(J=0\)} \\
0.99 & \text{if $A \oplus D = J-1$} \\
0 & \text{if $A \oplus D \neq J-1$} \\
\end{cases}
\]
we then assign different CPTs for \(S\):

\[\theta^1_{S | D, Y, J} = \begin{cases}
0.5 & \text{if \(J=0\)} \\
0.99 & \text{if $J \neq 0$ and \((J-1) \oplus D \oplus Y = S\)} \\
0.01 & \text{if $J \neq 0$ and \((J-1) \oplus D \oplus Y \neq S\)} \\
\end{cases}
\]

\[\theta^2_{S | D, Y, J} = \begin{cases}
0.5 & \text{if \(J=0\)} \\
0.99 & \text{if $J \neq 0$ and $S=0$} \\
0.01 & \text{if $J \neq 0$ and $S=1$} \\
\end{cases}
\]
The two parameterizations induce a same \(\Pr(\V)\) yet different \(\Pr_{Y=0}(J=1, S=1).\)
\end{proof}

\begin{proposition}
\label{prop:csi-var-useful}
There exists a tuple \(\langle  G, \V, \CC \rangle\) where $\CC$ contains only CSI constraints for which \(\Pr_\X(\Y)\) is unidentifiable wrt \(\langle G, \V \rangle\) but is identifiable wrt \(\langle  G, \V, \CC \rangle.\)
\end{proposition}
\begin{proof}
Consider the causal graph in Figure~\ref{sfig:csi-eg2} and the CSI \((J \indep D\ |\ A, Y=0).\) The classical causal effect \(\Pr_J(S)\) is unidentifiable by the \textsc{ID} algorithm~\citep{aaai/ShpitserP06} since the subgraph containing \(\{A, Y ,J\}\) forms a \emph{hedge} wrt the causal effect.\footnote{A \emph{hedge} is a graphical notion that characterizes the unidentifiability of causal effects; see~\citep{aaai/ShpitserP06} for details.} 

We next show that the causal effect is identifiable with the CSI constraint. By Proposition~\ref{prop:state-var-eq}, the variable-based causal effect \(\Pr_J(S)\) is identifiable iff the state-based causal effect \(\Pr_{J=0}(S=0)\) is identifiable since none of the contexts appear in the the CSI. The CI-graph under the context \(\{J=0, S=0\}\) is identical to the original causal graph, and its mutilated graph \(T\) is depicted in Figure~\ref{sfig:csi-eg-42}. The only c-component in the mutilated graph is \(\{S, D\}\) and its corresponding c-factor is \(Q[SD]_{J=0,S=0}.\) We now construct the context \(\s:\{Y=0\}\) with the corresponding CI-graph \(H\) shown in Figure~\ref{sfig:csi-eg-43}. Since \(\s\) and \(H\) satisfy the condition in line~7 of Algorithm~\ref{alg:id-csi}, \(Q[SD]_{J=0,S=0}\) is identifiable if \(Q[SD]_{Y=0}\) is identifiable by the \textsc{id-qfunc} procedure. This is indeed the case since the c-component \(\{S,D\}\) is a maximal c-component in \(H.\)
\end{proof}

\begin{figure}[tb]
\centering
\begin{subfigure}[b]{0.33\linewidth}
\centering
\begin{tikzpicture}[->=stealth,auto,scale=\dagsize,transform shape]
\node[state,font=\huge] (A) at (0,0) {$A$};
\node[font=\huge] (C) at (-1.5,-2) {$Y$};
\node[state,font=\huge] (B) at (3,0) {$D$};
\node[font=\huge] (D) at (1.5,-2) {$J$};
\node[font=\huge] (E) at (4.5,-2) {$S$};
\path (A) edge (C);
\path (A) edge (D);
\path (B) edge (D);
\path (B) edge (E);
\path (D) edge (E);
\path (C) edge (D);
\end{tikzpicture}
\caption{CI-graph~\ref{sfig:csi-eg2}}
\label{sfig:csi-eg-41}
\end{subfigure}
\begin{subfigure}[b]{0.33\linewidth}
\centering
\begin{tikzpicture}[->=stealth,auto,scale=\dagsize,transform shape]
\node[state,font=\huge] (B) at (3,0) {$D$};
\node[font=\huge] (E) at (4.5,-2) {$S$};
\path (B) edge (E);
\end{tikzpicture}
\caption{mutilated~\ref{sfig:csi-eg-41}}
\label{sfig:csi-eg-42}
\end{subfigure}
\begin{subfigure}[b]{0.32\linewidth}
\centering
\begin{tikzpicture}[->=stealth,auto,scale=\dagsize,transform shape]
\node[state,font=\huge] (A) at (0,0) {$A$};
\node[font=\huge] (C) at (-1.5,-2) {$Y$};
\node[state,font=\huge] (B) at (3,0) {$D$};
\node[font=\huge] (D) at (1.5,-2) {$J$};
\node[font=\huge] (E) at (4.5,-2) {$S$};
\path (A) edge (C);
\path (A) edge (D);
\path (B) edge (E);
\path (D) edge (E);
\path (C) edge (D);
\end{tikzpicture}
\caption{context \(Y=0\)}
\label{sfig:csi-eg-43}
\end{subfigure}
\caption{context-induced and mutilated graphs for the examples in Figure~\ref{fig:csi-eg} under the CSI \((J \indep D\ |\ A, Y=0).\) }
\label{fig:csi-eg4}
\end{figure} 

Before showing Proposition~\ref{prop:id-equiv}, we present the following Lemma~\ref{lem:extend-state} which will be used for the proof of Proposition~\ref{prop:id-equiv}.
\begin{lemma}
\label{lem:extend-state}
Let \(\CC_1\) and \(\CC_2\) be two sets of state constraints where \(\CC_1 \subseteq \CC_2.\)\footnote{We say \(\CC_1 \subseteq \CC_2\) if for each variable \(T\), the states of \(T\) in \(\CC_1\) are a subset of the states of \(T\) in \(\CC_2.\)} Then a causal effect \(\Pr_\x(\y)\) is identifiable wrt \(\langle G, \V, \CC_1 \rangle\) if it is identifiable wrt \(\langle G, \V, \CC_2 \rangle.\)
\end{lemma}

\begin{proof}
By induction, it suffices to show that an unidentifiable causal effect \(\Pr_\x(\y)\) remains unidentifiable if we add a new state to any variable \(W.\) Since \(\Pr_\x(\y)\) is unidentifiable, there exist two parameterizations \(\Theta_1, \Theta_2\) that induce a same \(\Pr(\V)\) but different values for \(\Pr_\x(\y).\) Our proof is based on constructing parameterizations \(\Theta'_1, \Theta'_2\) that contain an additional state for \(W\) while inducing unidentifiability.

Let \(\P = \{P_1, \dots, P_k\}\) and \(\C = \{C_1, \dots, C_t\}\) be the parents and children of \(W.\) Moreover, let \(w_1, \dots w_m\) be the current states and \(w'\) be the new state of \(W.\) We construct new parameterizations \(\Theta'_1, \Theta'_2\) from the original parameterizations \(\Theta_1, \Theta_2\) as follows. Let \(\epsilon\) be a positive constant in $(0,1)$. We modify the CPTs for \(W\) and each child \(C \in \C\):
\[\theta'_{w' | \p} = \theta_{w_1 | \p} \cdot \epsilon\ \ \ \ \theta'_{w_1 | \p} = \theta_{w_1 | \p} \cdot (1 - \epsilon)\]
\[\theta'_{c | \p_{C},w'} = \theta_{c | \p_{C}, w_1}\]
where \(\P_C\) denotes the parents of \(C\) excluding \(W.\) Let \(\Pr'_1\) and \(\Pr'_2\) be the distribution induced by \(\Theta'_1\) and \(\Theta'_2\). For each instantiation \((\u,\v)\) over hidden and observed variables,
\[\Pr'(\u,\v) = \begin{cases}
  \Pr((\u,\v)|_{w_1}) \cdot \epsilon  & w' \in (\u, \v) \\
  \Pr(\u,\v) \cdot (1-\epsilon) & w_1 \in (\u, \v) \\
  \Pr(\u,\v) & \text{otherwise}\\
\end{cases}
\]
where \((\u,\v)|_{w_1}\) replaces the state of \(W\) in \((\u,\v\)) by \(w_1.\)
Suppose \(W\) is hidden, summing out \(\U\) yields \(\Pr'(\V) = \Pr(\V)\); hence, \(\Pr'_1(\V) = \Pr_1(\V) = \Pr_2(\V) = \Pr'_2(\V).\) Suppose \(W\) is observed, we can compute \(\Pr'(\V)\) as follows:
\[\Pr'(\v) = \begin{cases}
  \Pr(\v|_{w_1}) \cdot \epsilon  & w' \in  \v \\
  \Pr(\v) \cdot (1-\epsilon) & w_1 \in \v \\
  \Pr(\v) & \text{otherwise}\\
\end{cases}
\]
Again, \(\Pr'_1(\V) = \Pr'_2(\V)\) since \(\Pr_1(\V) = \Pr_2(\V).\) 

We are left to show that \(\Pr'_{1\x}(\y) \neq \Pr'_{2\x}(\y).\) Suppose \(W \in \U,\) the distribution \(\Pr_\x(\V)\) is preserved and therefore \(\Pr'_{1\x}(\y) = \Pr_{1\x}(\y) \neq \Pr_{2\x}(\y) = \Pr'_{2\x}(\y).\) Suppose \(W \in \V,\) we can again compute \(\Pr'_\x(\y)\) as follows:
\[\Pr'_\x(\y) = \begin{cases}
  \Pr_\x(\y) & W \notin \Y\\
  \Pr_\x(\y|_{w_1}) \cdot \epsilon  & {W \in \Y} \text{ and } {w' \in \y} \\
  \Pr_\x(\y) \cdot (1-\epsilon) & {W \in \Y} \text{ and } {w_1 \in \y} \\
\end{cases}
\]
In all cases, \(\Pr'_{1\x}(\y)\neq \Pr'_{2\x}(\y)\).
\end{proof}
\begin{proposition}
\label{prop:id-equiv}
Let \(\CC\) be a set of state constraints and \(G\) be a Semi-Markovian graph. A causal effect \(\Pr_\X(\Y)\) (or \(\Pr_\x(\y)\)) is identifiable wrt \(\langle G, \V \rangle\) iff \(\Pr_\X(\Y)\) (or \(\Pr_\x(\y)\)) is constrained-identifiable wrt \(\langle G, \V, \CC \rangle\).
\end{proposition}
\begin{proof}
The only-if direction follows from the definition. We now prove the contrapositive of the if direction. According to the \textsc{id} algorithm~\citep{aaai/ShpitserP06}, \(\Pr_\X(\Y)\) is unidentifiable iff there exist two parameterizations \(\Theta^1\) and \(\Theta^2\) that induce a same \(\Pr(\V)\) yet different \(\Pr_\x(\y)\) and in which \emph{all variables are binary}. Hence, we can apply Lemma~\ref{lem:extend-state} to construct parameterizations with arbitrary numbers of variable states that induce unidentifiability. The equivalence between \(\Pr_\X(\Y)\) and \(\Pr_\x(\y)\) follows from the fact that we can always permute the states of \(\X\) and \(\Y.\)
\end{proof}

\begin{proposition}
\label{prop:cid-states}
There exist a causal graph \(G\), observed variables \(\V\), CSI constraints \(\CC_1\), and state constraints \(\CC_2\) for which \(\Pr_\X(\Y)\) is unidentifiable wrt \(\langle  G, \V, \CC_1 \rangle\) but is identifiable wrt \(\langle G, \V, \CC_1 \cup \CC_2 \rangle.\)
\end{proposition}
\begin{proof}
Consider the causal graph in Figure~\ref{sfig:state-eg1} and the CSI constraints (\(X \indep U | A=0\)) and (\(Y \indep U | X, A=1\)). We first show that the variable-based causal effect is unidentifiable without restricting variable states. Assume variables \(U, X, Y\) have states \(\{0, 1\}\) and \(A\) has states \(\{0, 1, 2\}\). We construct two parameterizations \(\Theta^1, \Theta^2\) that induce a same \(\Pr(\V)\) yet distinct answers for the causal effect \(\Pr_{X=0}(Y=0).\) In both parameterizations, we assign uniform distributions for \(U, A\) and the following CPT for \(X:\)

\[\theta^1_{X | U, A} = \theta^2_{X | U, A} = \begin{cases}
0.5 & \text{if \(A=0\) or \(A=1\)} \\
1 & \text{if $A = 2$ and $X=U$} \\
0 & \text{if $A = 2$ and $X \neq U$} \\
\end{cases}
\]

We then assign different CPTs for $Y$ as follows:
\[\theta^1_{Y | U, X, A} = \begin{cases}
0.5 & \text{if \(A=0\) or \(A=1\)} \\
0.99 & \text{if $A = 2$ and \(U \oplus X = Y\)} \\
0.01 & \text{if $A = 2$ and \(U \oplus X \neq Y\)} \\
\end{cases}
\]

\[\theta^2_{Y | U, X, A} = \begin{cases}
0.5 & \text{if \(A=0\) or \(A=1\)} \\
0.99 & \text{if $A = 2$ and \(Y=0\)} \\
0.01 & \text{if $A = 2$ and \(Y=1\)} \\
\end{cases}
\]

The two parameterizations induce a same \(\Pr(\V)\) yet distinct \(\Pr_{X=0}(Y)\) and \(\Pr_{X=1}(Y).\) 

We next apply Algorithm~\ref{alg:id-csi-state} to show that the causal effect \(\Pr_X(Y)\) is identifiable if \(A\) has states \(\{0, 1\}.\) We consider the context-partition \(\{A\Equal 0, A\Equal 1\}\) and reduce the problem of identifying \(\Pr_X(Y)\) to that of identifying sub-queries \(\Pr_{X=0}(Y=0, A=0)\) and \(\Pr_{X=0}(Y=0, A=1).\) Figures~\ref{sfig:state1-2} and~\ref{sfig:state1-3} depict the CI-graphs for the two sub-queries, and Figures~\ref{sfig:csi-eg-51} and~\ref{sfig:csi-eg-52} depict their mutilated graphs. 

To identify the two c-factors in Figures~\ref{sfig:csi-eg-51}, we consider the context \(A\Equal 0\) and its CI-graph \(T\) which coincides with Figure~\ref{sfig:state1-2}. Since \(\{U, Y\}\) and \(\{A\}\) are maximal c-components in \(T,\) their c-factors \(Q[UY]_{A=0}\) and \(Q[A]_{A=0}\) are identifiable. It follows that \(Q[UY]_{A=0,X=0,Y=0}\) and \(Q[A]_{A=0,X=0,Y=0}\) are also identifiable, which concludes the identifiability of \(\Pr_{X=0}(Y=0,A=0)\).

To identify the two c-factors in Figures~\ref{sfig:csi-eg-52}, we consider the context \(A\Equal 1\) and its CI-graph \(T\) which coincides with Figure~\ref{sfig:state1-3}. Since both \(\{Y\}\) and \(\{A\}\) are c-components in \(T,\) their corresponding c-factors \(Q[Y]_{A=1}\) and \(Q[A]_{A=1}\) are computable from \(\Pr(\V),\) which establishes the identifiability of \(\Pr_{X=0}(Y=0,A=1).\)
\end{proof}

\begin{figure}[tb]
\centering
\begin{subfigure}[b]{0.24\linewidth}
\centering
\begin{tikzpicture}[->=stealth,auto,scale=\dagsize,transform shape]
\node[state,font=\huge] (A) at (0,0) {$U$};
\node[font=\huge] (D) at (1.5,-1.5) {$Y$};
\node[font=\huge,red] (B) at (0,-3) {$A$};

\path (A) edge (D);
\path (B) edge (D);
\end{tikzpicture}
\caption{mutilated~\ref{sfig:state1-2}}
\label{sfig:csi-eg-51}
\end{subfigure}
\begin{subfigure}[b]{0.24\linewidth}
\centering
\begin{tikzpicture}[->=stealth,auto,scale=\dagsize,transform shape]
\node[font=\huge] (D) at (1.5,-1.5) {$Y$};
\node[font=\huge, red] (B) at (0,-3) {$A$};

\path (B) edge (D);
\end{tikzpicture}
\caption{mutilated~\ref{sfig:state1-3}}
\label{sfig:csi-eg-52}
\end{subfigure}
\begin{subfigure}[b]{0.24\linewidth}
\centering
\begin{tikzpicture}[->=stealth,auto,scale=\dagsize,transform shape]
\node[state,font=\huge] (B) at (2,0) {$U_2$};
\node[font=\huge,red] (D) at (4,0) {$B$};
\node[font=\huge] (F) at (2,-2) {$A$};
\node[font=\huge] (G) at (4,-2) {$Y$};

\path (B) edge (F);
\path (B) edge (G);
\path (D) edge (G);
\path (F) edge (G);
\end{tikzpicture}
\caption{mutilated~\ref{sfig:state2-2}}
\label{sfig:csi-eg-61}
\end{subfigure}
\begin{subfigure}[b]{0.24\linewidth}
\centering
\begin{tikzpicture}[->=stealth,auto,scale=\dagsize,transform shape]
\node[state,font=\huge] (B) at (2,0) {$U_2$};
\node[font=\huge,red] (D) at (4,0) {$B$};
\node[font=\huge] (G) at (4,-2) {$Y$};
\node[state,font=\huge] (C) at (2,-4) {$U_3$};

\path (B) edge (G);
\path (D) edge (G);
\path (C) edge (G);
\end{tikzpicture}
\caption{mutilated~\ref{sfig:state2-3}}
\label{sfig:csi-eg-62}
\end{subfigure}
\caption{mutilated graphs for the examples in Figure~\ref{fig:state-alg1}.}
\label{fig:csi-eg5}
\end{figure} 

\begin{proposition}
\label{prop:diff-domain}
There exist a causal graph \(G\), observed variables \(\V\), CSI constraints \(\CC_1\), state constraints \(\CC_2\), and particular states \(\x,\) and \(\y\) where \(\Pr_\x(\y)\) is unidentifiable wrt \(\langle  G, \V, \CC_1 \rangle\) but is identifiable wrt \(\langle  G,\) \( \V,\) \(\CC_1 \cup \CC_2 \rangle\). Moreover, \(\Pr_\X(\Y)\) is unidentifiable wrt \(\langle  G, \V, \CC_1 \cup \CC_2 \rangle\).
\end{proposition}
\begin{proof}
Consider the second example for Figure~\ref{sfig:state-eg2} in Section~\ref{sec:domain-constraints}. For simplicity, let \textit{raining} be the state $0$ and \textit{snowing} be the state $1$ of \(B\), and let \textit{low-cost} be the state $0$ of \(X.\) We can rewrite the CSIs as (\(A \indep U_1\ |\ X = 0, U_2\)), (\(Y \indep U_3\ |\ B=0, U_2, A\)), and (\(Y \indep A | B=1, U_2, U_3\)).

We first show \(\Pr_{X=0}(Y=0)\) is unidentifiable by finding two parameterizations \(\Theta^1, \Theta^2\) that induce a same \(\Pr(\V)\) yet different \(\Pr_{X=0}(Y=0)\). We assume that all variables have states \(\{0,1\}\), except for \(B\), which has states \(\{0, 1, 2\}.\) For both parameterizations, we assign uniform distributions for \(U_1, U_2, U_3\), structural equations \(X = U_3,\) \(A = U_2 \oplus X,\) and the following CPT for \(B\):
\[\theta^1_B = \theta^2_B = \begin{cases}
  0.05 & \text{if $B=0$ or $B=1$}\\
  0.9 & \text{if $B=2$}\\
\end{cases}
\]
We assign different CPTs for \(Y\) in \(\Theta_1\) and \(\Theta_2\):
\[\theta^1_{Y | U_2, U_3, A, B} = \begin{cases}
  0.5 & \text{if $B=0$ or $B=1$}\\
  0.99 & \text{if $B=1$ and $Y = U_2 \oplus A \oplus U_3$}\\
  0.01 & \text{if $B=1$ and $Y \neq U_2 \oplus A \oplus U_3$}\\
\end{cases}
\]
\[\theta^2_{Y | U_2, U_3, A, B} = \begin{cases}
  0.5 & \text{if $B=0$ or $B=1$}\\
  0.99 & \text{if $B=1$ and $Y=0$}\\
  0.01 & \text{if $B=1$ and $Y=1$}\\
\end{cases}
\]
The two parameterizations attains a same \(\Pr(\V)\) yet different answers for \(\Pr_{X=0}(Y=0).\)

We now apply Algorithm~\ref{alg:id-csi-state} to show that \(\Pr_{X=0}(Y=0)\) is identifiable if the states of \(B\) are restricted to \(\{0,1\}\). We consider the context-partition \(\{B\Equal 0, B\Equal 1\}\) and show that the sub-queries \(\Pr_{X=0}(Y=0, B=0)\) and \(\Pr_{X=0}(Y=0, B=1)\) are identifiable. Figures~\ref{sfig:state2-2} and~\ref{sfig:state2-3} depict the CI-graphs for the sub-queries, and Figures~\ref{sfig:csi-eg-61} and~\ref{sfig:csi-eg-62} depict their mutilated graphs.

We first identify the c-factors of c-components \(\{U_2, A, Y\}\) and \(\{B\}\) in the mutilated graph for sub-query \(\Pr_{X=0}(Y=0, B=0)\). Consider the context \(\s: \{X\Equal 0, B\Equal 0\}\) and its CI-graph \(H\) which coincides with Figure~\ref{sfig:csi-eg-61} (Algorithm~\ref{alg:id-csi} lines~6-7). Since both \(\{U_2, A, Y\}\) and \(\{B\}\) are maximal c-components in \(H,\) \(\Pr_{X=0}(Y=0, B=0)\) is identifiable. 

We now identify the c-factors of c-components \(\{U_2, U_3, Y\}\) and \(\{B\}\) in the mutilate graph for sub-query \(\Pr_{X=0}(Y=0, B=1).\) We consider the context \(\s: \{B\Equal 1\}\) and its CI-graph \(H\) shown in Figure~\ref{fig:csi-eg-64}. The first c-component can be obtained from \(H\) by first pruning leaves \(A, X\) and then taking the maximal c-component. The second c-component is a maximal c-component of \(H.\) Hence, \(\Pr_{X=0}(Y=0, B=1)\) is identifiable by the soundness of Algorithm~\ref{alg:id-csi}.
\begin{figure}[tb]
\centering
\begin{tikzpicture}[->=stealth,auto,scale=\dagsize,transform shape]
\node[state,font=\huge] (A) at (0,0) {$U_1$};
\node[state,font=\huge] (B) at (2,0) {$U_2$};
\node[font=\huge] (D) at (4,0) {$B$};
\node[font=\huge] (E) at (0,-2) {$X$};
\node[font=\huge] (F) at (2,-2) {$A$};
\node[font=\huge] (G) at (4,-2) {$Y$};
\node[state,font=\huge] (C) at (2,-4) {$U_3$};

\path (A) edge (E);
\path (A) edge (F);
\path (B) edge (F);
\path (B) edge (G);
\path (D) edge (G);
\path (E) edge (F);
\path (C) edge (E);
\path (C) edge (G);
\end{tikzpicture}
\caption{CI-graph of Figure~\ref{sfig:state-eg2} under the context \(B\Equal 1\).}
\label{fig:csi-eg-64}
\end{figure}

We finally show that \(\Pr_X(Y)\) is unidentifiable under both the CSI and state constraints by constructing parameterizations \(\Theta^1, \Theta^2\) that form an instance of unidentifiability. We assume all variables are binary with states \(\{0,1\}\), except for \(X\), which has states \(\{0, 1, 2\}.\) For both \(\Theta^1\) and \(\Theta^2\), we assign uniform distributions for variables \(U_1, U_2, U_3, B\) and the following CPTs for \(X, Y\):
\[\theta^1_{X | U_1, U_3} = \theta^2_{X | U_1, U_3} = \begin{cases}
  (0.05, 0.95, 0) & \text{if $U_1=0$}\\
  (0.05, 0, 0.95) & \text{if $U_1=1$}\\
\end{cases}
\]
\[\theta^1_{Y | U_2, U_3, A, B} =  \theta^2_{Y | U_2, U_3, A, B} = \begin{cases}
  0.99 & \text{if $A=Y$}\\
  0.01 & \text{if $A \neq Y$}\\
\end{cases}
\]
We assign distinct CPTs for \(A\) in \(\Theta^1\) and \(\Theta^2\):
\[\theta^1_{A | U_1, U_2, X} = \begin{cases}
  0.5 & \text{if $X=0$}\\
  0.99 & \text{if $X \neq 0$ and $(X-1) \oplus U_1 = A$}\\
  0.01 & \text{if $X \neq 0$ and $(X-1) \oplus U_1 \neq A$}\\
\end{cases}
\]
\[\theta^2_{A | U_1, U_2, X} = \begin{cases}
  0.5 & \text{if $X=0$}\\
  0.99 & \text{if $X \neq 0$ and $A = 0$}\\
  0.01 & \text{if $X \neq 0$ and $A = 1$}\\
\end{cases}
\]
The parameterizations induce a same \(\Pr(\V)\) yet different values for \(\Pr_{X=1}(Y=1).\)
\end{proof}

\begin{proposition}
\label{prop:state-var-eq-ext}
Let \(\CC_1\) be a set of CSI constraints and \(\CC_2\) be a set of state constraints. Suppose \(\CC_2\) mentions treatment variables \(\X_1\) and outcome variables \(\Y_1\). Let \(\X_2 = \X \setminus \X_1\) and \(\Y_2 = \Y \setminus \Y_1.\) Then \(\Pr_{\X}(\Y)\) is identifiable (wrt \(\langle G, \V, \CC_1 \cup \CC_2 \rangle\)) iff \(\Pr_{\x_1\x_2}(\y_1\y_2)\) is identifiable for all assignments \(\x_1,\y_1\) specified by \(\CC_2\) and any assignment \(\x_2,\y_2\) that do not appear in the contexts of \(\CC_1.\)
\end{proposition}
\begin{proof}
The only-if direction follows from the definition of variable-based identifiability (Definition~\ref{def:constrained-id}). We now prove the contrapositive of the if direction. Suppose \(\Pr_\X(\Y)\) is unidentifiable, there exist two parameterizations that induce a same \(\Pr(\V)\) but different \(\Pr_{\x_1\x_2}(\y_1\y_2)\), where \(\x_1,\y_1\) are states specified by \(\CC_2.\) 
We then iterate over each state \(s\) in \(\x_2\y_2\) and apply the construction method from the proof of Proposition~\ref{prop:state-var-eq}. The resulting parameterizations satisfy \(\CC_1\) and \(\CC_2\), induce a same \(\Pr(\V)\), and yield different values of \(\Pr_{\x_1\x'_2}(\y_1\y'_2)\) for \(\x'_2,\y'_2\) outside the CSI contexts. 
\end{proof}

\begin{corollary}
\label{cor:state-var-eq-ext}
Let \(\CC_1\) be a set of CSI constraints and \(\CC_2\) be a set of state constraints. If the states \(\x,\y\) do not appear in the context of \(\CC_1\) and variables \(\X, \Y\) are not mentioned in \(\CC_2\), then \(\Pr_{\X}(\Y)\) is identifiable wrt \(\langle G, \V, \CC_1 \cup \CC_2 \rangle\) iff \(\Pr_{\x}(\y)\) is identifiable wrt \(\langle G, \V, \CC_1 \cup \CC_2 \rangle\).
\end{corollary}
\begin{proof}
Since \(\CC_2\) does not mention variables \(\X\) and \(\Y\), the variable sets in Proposition~\ref{prop:state-var-eq-ext} take the following values: \(\X_1 = \emptyset\), \(\Y_1 = \emptyset,\) \(\X_2 = \X\), and \(\Y_2 = \Y\). Hence, \(\Pr_\X(\Y)\) is identifiable iff \(\Pr_\x(\y)\) is identifiable for any \(\x,\y\) not appearing in the CSI contexts.
\end{proof}

\begin{table*}[tb]
\renewcommand{\arraystretch}{1.25} 
\scriptsize
\centering
\begin{tabular}{|c|c|c||c||c|c||c|c|c|}
\hline
\multirow{2}{*}{$n$}  & \multirow{2}{*}{$r$} & \multirow{2}{*}{stats} & \textsc{classical-id} & \multicolumn{2}{c||}{${\textsc{id-csi-heu}}$}  & \multicolumn{2}{c|}{${\textsc{id-csi-state-heu}}$} & \multicolumn{1}{c|}{${\textsc{dosearch}}$}\\
\cline{4-9}
 & & & {variable \& state} & variable & state & variable & state & variable \\
\hline
\multirow{6}{*}{6} & \multirow{2}{*}{0.2} & \#ID/success & 171/200 & 171/200 & 171/200 & 171/200 & 171/200 & 171/200 \\
\cline{3-9}
& & runtime (s) & 0.001 & 0.001 & 0.001 & 0.002 & 0.002 & 0.9 \\
\cline{2-9}
& \multirow{2}{*}{0.5} & \#ID/success & 171/200 & 172/200 & 174/200 & 173/200 & 176/200 & 173/200 \\
\cline{3-9}
& & runtime (s) & 0.001 & 0.001 & 0.002 & 0.002 & 0.002 & 0.9\\
\cline{2-9}
& \multirow{2}{*}{0.8} & \#ID/success & 171/200 & 171/200 & 172/200 & 175/200 & 176/200 & 174/200\\
\cline{3-9}
& & runtime (s) & 0.001 & 0.001 & 0.001 & 0.002 & 0.002 & 0.9\\
\hline 
\hline
\multirow{8}{*}{8} & \multirow{2}{*}{0.2} & \#ID/success & 120/200 & 120/200 & 123/200 & 121/200 & 124/200 & 120/200\\
\cline{3-9}
& & runtime (s) & 0.002 & 0.002 & 0.003 & 0.003 & 0.002 & 6.2\\
\cline{2-9}
& \multirow{2}{*}{0.5} & \#ID/success & 120/200 & 121/200 & 132/200 & 123/200 & 133/200 & 122/199\\
\cline{3-9}
& & runtime (s) & 0.002 & 0.002 & 0.002 & 0.005 & 0.003 & 4.9\\
\cline{2-9}
& \multirow{2}{*}{0.8} & \#ID/success & 120/200 & 121/200 & 133/200 & 126/200 & 137/200 & 123/198 \\
\cline{3-9}
& & runtime (s) & 0.002 & 0.002 & 0.002 & 0.005 & 0.003 & 6.6\\
\hline 
\hline
\multirow{6}{*}{10} & \multirow{2}{*}{0.2} & \#ID/success & 115/200 & 115/200 & 118/200 & 115/200 & 118/200 & 112/198\\
\cline{3-9}
& & runtime (s) & 0.003 & 0.004 & 0.003 & 0.004 & 0.003 & 14.0\\
\cline{2-9}
& \multirow{2}{*}{0.5} & \#ID/success & 115/200 & 117/200 & 127/200 & 120/200 & 132/200 & 116/191\\
\cline{3-9}
& & runtime (s) & 0.002 & 0.003 & 0.003 & 0.005 & 0.005 & 42.5 \\
\cline{2-9}
& \multirow{2}{*}{0.8} & \#ID/success & 115/200 & 119/200 & 132/200 & 129/200 & 140/200 & 124/190\\
\cline{3-9}
& & runtime (s) & 0.002 & 0.003 & 0.003 & 0.007 & 0.004 & 40.2\\
\hline 
\hline
\multirow{6}{*}{12} & \multirow{2}{*}{0.2} & \#ID/success & 107/200 & 108/200 & 113/200 & 109/200 & 114/200 & 99/175\\
\cline{3-9}
& & runtime (s) & 0.003 & 0.003 & 0.004 & 0.005 & 0.004 & 92.2\\
\cline{2-9}
& \multirow{2}{*}{0.5} & \#ID/success & 107/200 & 109/200 & 121/200 & 114/200 & 124/200 & 79/122 \\
\cline{3-9}
& & runtime (s) & 0.003 & 0.003 & 0.003 & 0.007 & 0.005 & 165.3 \\
\cline{2-9}
& \multirow{2}{*}{0.8} & \#ID/success & 107/200 & 109/200 & 131/200 & 121/200 & 136/200 & 79/127\\
\cline{3-9}
& & runtime (s) & 0.003 & 0.003 & 0.003 & 0.009 & 0.005 & 152.0\\
\hline 
\hline
\multirow{6}{*}{15} & \multirow{2}{*}{0.2} & \#ID/success & 84/200 & 84/200 & 90/200 & 85/200 & 90/200 & - \\
\cline{3-9}
& & runtime (s) & 0.004 & 0.005 & 0.004 & 0.007 & 0.006 & - \\
\cline{2-9}
& \multirow{2}{*}{0.5} & \#ID/success & 84/200 & 84/200 & 101/200 & 87/200 & 104/200 & - \\
\cline{3-9}
& & runtime (s) & 0.004 & 0.004 & 0.005 & 0.01 & 0.01 & - \\
\cline{2-9}
& \multirow{2}{*}{0.8} & \#ID/success & 84/200 & 85/200 & 108/200 & 89/200 & 110/200 & - \\
\cline{3-9}
& & runtime (s) & 0.004 & 0.004 & 0.004 & 0.02 & 0.01 & - \\
\hline 
\hline
\multirow{6}{*}{20} & \multirow{2}{*}{0.2} & \#ID/success & 69/200 & 69/200 & 72/200 & 70/200 & 72/200 & - \\
\cline{3-9}
& & runtime (s) & 0.007 & 0.007 & 0.007 & 0.01 & 0.008 & - \\
\cline{2-9}
& \multirow{2}{*}{0.5} & \#ID/success & 69/200 & 72/200 & 86/200 & 78/200 & 90/200 & - \\
\cline{3-9}
& & runtime (s) & 0.007 & 0.007 & 0.007 & 0.03 & 0.01 & - \\
\cline{2-9}
& \multirow{2}{*}{0.8} & \#ID/success & 69/200 & 76/200 & 95/200 & 84/200 & 100/200 & - \\
\cline{3-9}
& & runtime (s) & 0.003 & 0.003 & 0.003 & 0.14 & 0.02 & - \\
\hline 
\hline
\multirow{6}{*}{30} & \multirow{2}{*}{0.2} & \#ID/success & 45/200 & 45/200 & 47/200 & 45/200 & 47/200 & - \\
\cline{3-9}
& & runtime (s) & 0.03 & 0.03 & 0.03 & 0.05 & 0.03 & - \\
\cline{2-9}
& \multirow{2}{*}{0.5} & \#ID/success & 45/200 & 48/200 & 61/200 & 51/200 & 62/200 & - \\
\cline{3-9}
& & runtime (s) & 0.03 & 0.03 & 0.03 & 0.17 & 0.05 & - \\
\cline{2-9}
& \multirow{2}{*}{0.8} & \#ID/success & 45/200 & 48/200 & 65/200 & 51/200 & 69/200 & - \\
\cline{3-9}
& & runtime (s) & 0.03 & 0.03 & 0.03 & 0.31 & 0.07 & - \\
\hline 
\hline
\multirow{6}{*}{40} & \multirow{2}{*}{0.2} & \#ID/success & 20/200 & 21/200 & 24/200 & 22/200 & 25/200 & - \\
\cline{3-9}
& & runtime (s) & 0.03 & 0.03 & 0.03 & 0.06 & 0.04 & - \\
\cline{2-9}
& \multirow{2}{*}{0.5} & \#ID/success & 20/200 & 22/200 & 30/200 & 22/200 & 30/200 & - \\
\cline{3-9}
& & runtime (s) & 0.03 & 0.03 & 0.03 & 0.19 & 0.06 & - \\
\cline{2-9}
& \multirow{2}{*}{0.8} & \#ID/success & 20/200 & 22/200 & 41/200 & 28/200 & 44/200 & - \\
\cline{3-9}
& & runtime (s) & 0.03 & 0.03 & 0.03 & 0.70 & 0.09 & - \\

\hline 
\hline
\multirow{6}{*}{50} & \multirow{2}{*}{0.2} & \#ID/success & 15/200 & 15/200 & 19/200 & 15/200 & 19/200 & - \\
\cline{3-9}
& & runtime (s) & 0.06 & 0.06 & 0.06 & 0.24 & 0.08 & - \\
\cline{2-9}
& \multirow{2}{*}{0.5} & \#ID/success & 15/200 & 19/200 & 31/200 & 19/199 & 33/200 & - \\
\cline{3-9}
& & runtime (s) & 0.06 & 0.07 & 0.06 & 2.68 & 0.21 & - \\
\cline{2-9}
& \multirow{2}{*}{0.8} & \#ID/success & 15/200 & 20/200 & 37/200 & 25/200 & 41/200 & - \\
\cline{3-9}
& & runtime (s) & 0.06 & 0.07 & 0.06 & 4.08 & 0.37 & - \\
\hline 

\end{tabular}
\caption{We record the number of variable-based (column ``variable'') and state-based (column ``state'') causal effects deemed identifiable for each method under causal graphs with varying numbers of nodes ($n$) and CSIs ($r$). We also record the number of runs that complete within the 5-minute timeout (labeled as ``success'' in the table). For runs that exceed the timeout, we set the runtime to 5 minutes when computing the average.}
\label{tab:exp1}
\end{table*}

\end{document}